\pdfoutput=1
\documentclass[11pt]{article}

\usepackage[preprint]{acl}

\usepackage{times}
\usepackage{latexsym}

\usepackage[T1]{fontenc}
\usepackage[utf8]{inputenc}

\usepackage{microtype}

\usepackage{inconsolata}

\usepackage{graphicx}
\usepackage{subcaption}
\usepackage{amsmath}
\usepackage{fancyvrb}
\usepackage{multicol}
\usepackage{booktabs}

\def\soulspeak{\textsc{SoulSpeak}}

\title{Advancing Conversational Psychotherapy: Integrating Privacy, Dual-Memory, and Domain Expertise with Large Language Models}

\author{XiuYu Zhang\thanks{Correspondence to xiuyuzhang@berkeley.edu} \\
University of California, Berkeley \\
 \And
  Zening Luo \\
University of California, Berkeley \\}

\begin{document}
\maketitle
\begin{abstract}
Mental health has increasingly become a global issue that reveals the limitations of traditional conversational psychotherapy, constrained by location, time, expense, and privacy concerns. In response to these challenges, we introduce \soulspeak{}, a Large Language Model (LLM)-enabled chatbot designed to democratize access to psychotherapy. \soulspeak{} improves upon the capabilities of standard LLM-enabled chatbots by incorporating a novel dual-memory component that combines short-term and long-term context via Retrieval Augmented Generation (RAG) to offer personalized responses while ensuring the preservation of user privacy and intimacy through a dedicated privacy module. In addition, it leverages a counseling chat dataset of therapist-client interactions and various prompting techniques to align the generated responses with psychotherapeutic methods. We introduce two fine-tuned BERT models to evaluate the system against existing LLMs and human therapists: the Conversational Psychotherapy Preference Model (CPPM) to simulate human preference among responses and another to assess response relevance to user input. CPPM is useful for training and evaluating psychotherapy-focused language models independent from \soulspeak{}, helping with the constrained resources available for psychotherapy. Furthermore, the effectiveness of the dual-memory component and the robustness of the privacy module are also examined. Our findings highlight the potential and challenge of enhancing mental health care by offering an alternative that combines the expertise of traditional therapy with the advantages of LLMs, providing a promising way to address the accessibility and personalization gap in current mental health services.

\end{abstract}

\section{Introduction}
Mental health illness has always been a global issue~\citep{wto-mental-health2022}. In an era of global economic uncertainty, rampant war, social isolation, ever-increasing class disparities, and general political nihilism, it is of little surprise that rates of depression and suicide have soared in recent years. Global rates of anxiety and depression increased by roughly 25\% within the first year of the COVID-19 pandemic~\citep{WorldHealthOrganization2022}. This increase in mental health issues has severe negative implications, including widespread suicide and decreased work productivity. Every year, according to a \citet{WorldHealthOrganization2023} press statement, over 700,000 people worldwide commit suicide, with countless more attempting to take their own life. It was estimated that poor mental health cost the world economy \$2.5 trillion USD per year in 2010, a figure predicted to rise to a staggering \$6 trillion USD per year by 2030~\citep{Lancet2020}. Psychotherapy can be used to treat many cases of mental health conditions~\citep{wto-depression2023}. Despite the pressing need for mental health care, the cost of therapy and psychiatric care remains prohibitively expensive for many. In one nationally representative study conducted in the United States by \citet{MentalHealthAmerica2023}, it was found that about 42\% of adults with any type of mental illness reported that they were unable to receive treatment because they could not afford it, highlighting the inaccessibility of mental health care to the general public. 

With the advancement of data-driven language models (LMs) in the field of Natural Language Processing (NLP), many works were developed to enhance mental health treatment, such as comparing the effectiveness of different therapy methods~\citep{Althoff2016LargescaleAO}, measuring the impact of the social medium on mental conditions~\citep{Choudhury2014MentalHD}, analyzing records of past treatments~\citep{Ive2020GenerationAE}, predicting mental conditions~\citep{Zhang2022NaturalLP}, etc. However, using LMs to generate psychotherapeutic responses directly remained challenging until the emergence of Large Language Models (LLMs). LLM-powered chatbots offer the tantalizing possibility of a low-cost, readily accessible alternative to traditional psychotherapy~\citep{He2023TowardsAP}. However, current efforts towards the development of therapeutic chatbots are often focused on immediate, short-term solutions, despite the fact that those who struggle with mental health issues may reap greater benefits from long-term care~\citep{Guo2024LargeLM}. Many therapy chatbots fail to remember critical user information from past conversations, greatly reducing their efficacy. This, combined with data privacy concerns, led to our proposal of \soulspeak{}, an LLM-based conversational agent with long-term memory, data anonymization, and domain expertise that provides effective, low-cost talk therapy to its users. 

\soulspeak{}, as detailed in Section~\ref{sec:method}, has several key advantages over existing online chatbots like ChatGPT. The first of which is its memory module, which compartmentalizes important information from conversations into short-term and long-term contexts. The memory module enables \soulspeak{} to keep track of crucial information from past conversations, such as important individuals and events, over long periods, greatly enhancing \soulspeak{}’s ability to understand, connect, and empathize with its users, a skill which is paramount in psychotherapy. Secondly, \soulspeak{} integrates an extensive dataset of anonymized conversations between professional therapists (Counsel Chat dataset)~\citep{bertagnolli2020counsel} and their clients into its Retrieval-Augmented Generation (RAG) process~\citep{NEURIPS2020_6b493230}, allowing it to provide responses that follow psychotherapeutic methods. Lastly, \soulspeak{} has a robust privacy module that utilizes a capable Named Entity Recognition (NER) model to detect, anonymize, and restore Personal Identifiable Information (PII), ensuring maximum user privacy. 

In addition, we introduce the Conversational Psychotherapy Preference Model (CPPM) to simulate human preference over responses to mental health-related queries. CPPM simulates human preference in responses to a therapeutic question with validation accuracy of over $97\%$. Together with other metrics to evaluate individual responses, CPPM is explained in Section~\ref{sec:individual}. We also discuss the statistical difference between human-produced and LLM-generated text in response to therapeutic questions, which can be used for future development. Furthermore, in Section~\ref{sec:dependent}, we test the effectiveness of \soulspeak{}’s long-term memory module and discuss the robustness of the privacy module. We conclude with a discussion of the limitations of this research and possible areas for future work in Section~\ref{sec:limitation}. 
\section{Background}\label{sec:background}
\subsection{Generative language models}

Language generation is an essential component of Artificial Intelligence (AI) and at the core of NLP. Alan Turing defined AI as a machine with the ability to exhibit intelligent behavior equivalent to, or indistinguishable from, that of a human. In his famous Turing test, intelligent behavior is accessed through a conversation where the AI generates text in response to human input. Early works on NLP approached language generation by manually setting linguistic rules and then shifted focus to statistical approaches. Statistical N-gram models~\citep{4766902} outperformed rule-based linguistic models and provided the foundation for Machine Learning (ML) based language models. Various data-driven models emerged with the rapid development of ML theories and computing devices, including the auto-regressive models developed on N-gram models to generate words one at a time based on previous sequential input or surrounding context. Due to the success of the Transformer model~\citep{Vaswani2017AttentionIA}, auto-regressive models dominated the NLP research afterward. Benefiting from the rise of deep learning and big data, large-scale generative language models such as Generative Pretrained Transformer (GPT)~\citep{Radford_Narasimhan_Salimans_Sutskever} and Bidirectional Encoder Representations from Transformers (BERT)~\citep{Devlin2019BERTPO} were created and implemented, exhibiting far superior performance than previous language models. GPT by OpenAI evolves over time with increased parameters from 117 million, 1.5 billion to 175 billion. The newest GPT model, GPT-4~\citep{Achiam2023GPT4TR}, can perform various types of text generation with multiple languages, showing the ability of in-context learning and chain of thought due to emergence.

\begin{figure*}[t]
\centering
\includegraphics[width=0.85\textwidth]{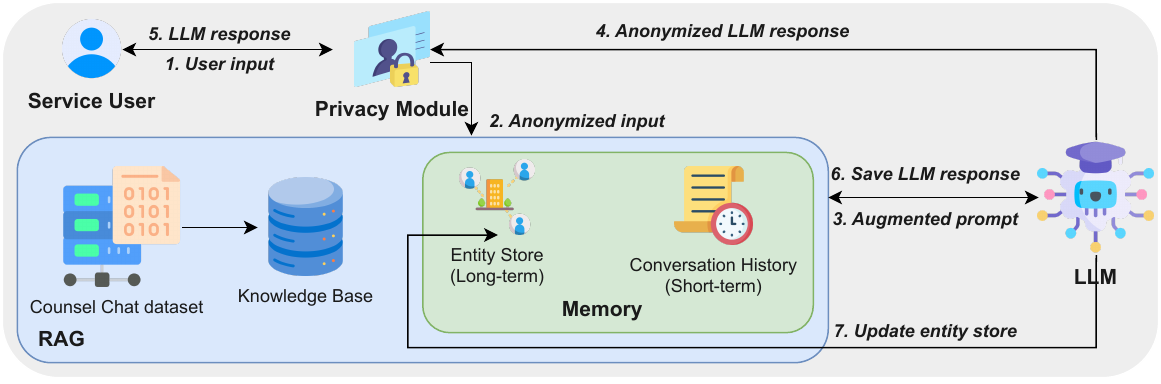}
\caption{The \soulspeak{} system architecture and user workflow.}
\label{fig:overview}
\end{figure*}
 
\subsection{Evolution of conversational psychotherapy: in-personal, online, chatbot}

Conversational psychotherapy is a common and effective method mental health practitioners use to work alongside their patients or service users~\citep{Psychotherapy-book}. In this paper and the scope of chatbots, the term service user is used as language model-powered agents are not licensed mental practitioners who can give medical treatment despite the ongoing debate over the usage of patient versus service users~\citep{priebe_2021}. Through conversational psychotherapy, mental health therapists build a personal and safe environment professionally for service users to express their thoughts and feelings~\citep{Swift2018TheIO}, requiring the therapists to emphatically interpret the words of the service users and respond in a carefully designed way to drive the conversation. However, traditional in-person psychotherapy requires the service user to make appointments in advance and go to a specific location at a specific time for a fixed period. This process is tedious, and the limitation of time and location may not work for everyone. Due to this, online psychotherapy quickly gained the favor of service users, which is more flexible, accessible, and affordable~\citep{Stoll2020EthicalII}. Despite the advantages of online psychotherapy over traditional psychotherapy, the availability of therapists and expenses remain an obstacle for some potential service users. 

With the advancement of generative LMs, chatbots become more capable of interpreting, modeling, and generating natural languages, which can be utilized for conversational psychotherapy~\citep{Denecke2020AMH, Chung2020ACF, He2022MentalHC, Wei2023LeveragingLL, Chen2023LLMempoweredCF}. As shown by \citet{Guo2024LargeLM} in their survey, there was a huge increase in conversational agents in 2022. This increase was a collaborative result of the increase in demand for psychotherapy due to COVID-19 and the emergence of LLMs. A similar study conducted by \citet{Romael2023} surveyed ten of the most prominent chatbot applications in the mental health space. They found that users generally valued the chatbots for their ability to communicate in a human-like manner and 24/7 availability; however, they also identified several key faults in popular chatbot technology. These included the usage of an overly cheerful tone in inappropriate situations and an overreliance on scripted responses. Moreover, the chatbots lacked the ability to reference previous conversations in new responses, instead relying on users to self-report mentioned people or events repeatedly. Lastly, the chatbots in the survey lacked the ability to recommend specific treatment plans based on individual user needs, offering only limited or short-term interventions. 

\section{Method}\label{sec:method}

Figure~\ref{fig:overview} presents the architecture of \soulspeak{}, our novel system designed to enhance the online therapeutic experience. The system comprises three main modules: the privacy module, the memory module, and the knowledge base with therapist domain expertise, all orchestrated around LLM APIs. We break down the overview in the following subsections. 

\subsection{Privacy module}
The privacy module is a gateway between the users and all the internal downstream processes in \soulspeak{}. Its primary role is to ensure that no PIIs are retained within the system's memory module nor transmitted to third-party APIs. We use a robust NER model adept at pinpointing PII, including person names, locations, and date times within user input. Upon detection, the identified PII is substituted with anonymized placeholders—randomly generated values that preserve the semantic structure without revealing sensitive information. These values are paired with their corresponding PII in a secure mapping to restore the original data into the LLM-generated content that references these anonymized entities. This process is partly inspired by the methodologies of Langchain \cite{langchain}. We further improve its implementation by integrating the PII detection process with our long-term memory module, which will be addressed in the following sections.

We use the model "en\_core\_web\_lg-3.7.1" from the spaCy NLP library as the default model to perform name entity recognition on the user input, which achieves a precision of 85.16\% on the benchmark dataset available from spaCy's model releases \citep{spacy_en_core_web_lg_3.7.1}.

\subsection{Knowledge base}

The knowledge base is designed to augment the user query to the LLM with domain expertise from certified therapists. To do so, we use the Counsel Chat dataset \cite{bertagnolli2020counsel}, a high-quality open-source dataset of counseling responses. The data consists of user questions, responses from licensed therapists, and metadata, including views and upvotes of the therapist's response, all extracted from counselchat.com.  A detailed introduction of the Counsel Chat dataset can be found in Appendix~\ref{appx:dataset}. We use a PostgreSQL instance with a vector extension to vectorize the questions into embeddings. Then, the anonymized user query will retrieve the most relevant question through a vector similarity search. Among all the responses to the question, the knowledge base will return the top-$k$ responses based on the preference score, as defined in Section \ref{subsec:preference}. In our experiments, we use $k=1$.

\subsection{Memory module}
The memory module consists of both short-term and long-term memory. The short-term memory maintains the most recent $n$ interactions between the user and LLM in its conversation history. During the RAG process, the agent appends the history to its prompt to provide the LLM with immediate context, allowing the agent to maintain a coherent and responsive dialogue flow. In our experiments, we set the $n=10$ by default. 

The long-term memory persists in LLM-generated summaries of various entities the user mentioned across different therapy sessions. It's realized through \textit{Entity Store}, which maps the entity names and their associated summaries. In particular, we leverage the PII detection infrastructure from the privacy module to insert entities into the entity store automatically. We recognize that the full context surrounding an entity typically spans several interactions between the user and the LLM. For example, when a user mentions a name during therapy sessions, the relevant details and context about that person are often scattered across multiple interactions. To ensure accuracy and completeness, we update the entity store periodically every $n$ interaction. This avoids updating the entity store too frequently with the same information and prevents potential overlaps between the entity summary and short-term memory, i.e., the recent conversation history.

When the Entity Store is updated, the existing entity summary and recent conversation history are sent to the LLM, which is instructed to update the existing summary only if new information is present; otherwise, it remains unchanged. When a user query mentions an entity, the Entity Store provides the most recent and relevant summary of that entity.

Integrating the two memory systems ensures that \soulspeak{} can make informed responses based on the immediate context and the broader narrative of the user's experiences. 

\subsection{Retrieval augmented generation process}
The final RAG process in \soulspeak{} integrates all previously mentioned components - the anonymized user query, retrieved therapist response from the knowledge base, short-term memory, and long-term memory - using the selected prompt template. Six different designs of prompts are implemented, which are discussed in detail in Appendix \ref{appendix:prompt}.

Note that the inclusion of retrieved therapist responses and long-term memory is conditional: the therapist response is incorporated only if the cosine similarity score between the user query and the questions in the therapy dataset exceeds a certain threshold, $\alpha$, and the long-term memory is included only if the user's query contains at least one entity name present in our entity store. In our experiments, we chose $\alpha = 0.2$. We conducted preliminary runs using values from 0.1 to 0.5, with increments of 0.1. We analyzed the retrieved therapy responses regarding relevance and variety and identified 0.2 as the sweet spot.

This approach guarantees that the information integrated into our final prompt is relevant and beneficial for the agent to respond to the user's needs.

\begin{figure*}[t]
\centering
\includegraphics[width=0.85\textwidth]{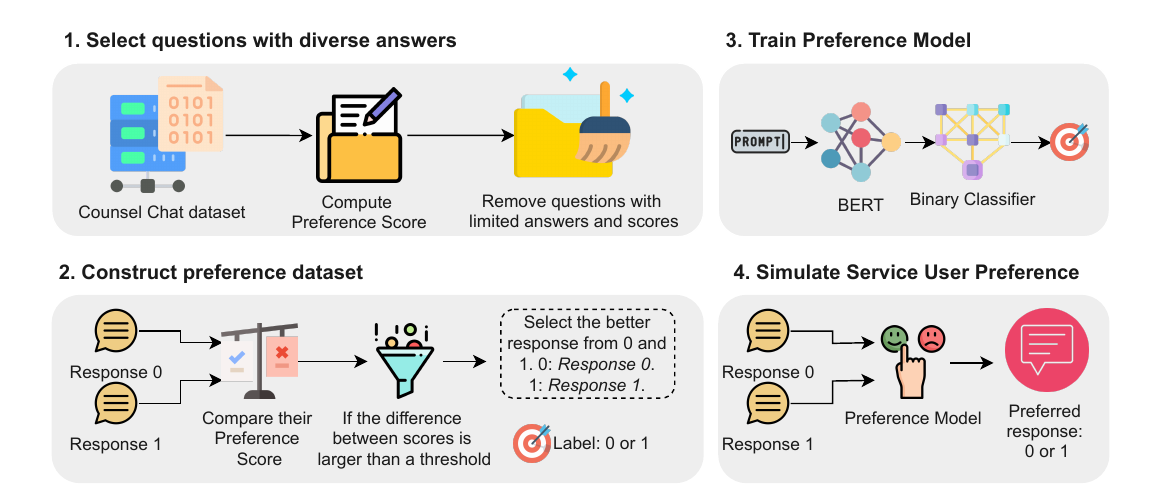}
\caption{Overview of the Conversational Psychotherapy Preference Model (CPPM) to simulate service users' preference over responses. Note that the two responses as input to the model are assumed to be for the same question or query.}
\label{fig:preference-model}
\end{figure*}

\section{Individual Response Analysis}\label{sec:individual}
This section examines \soulspeak{}'s performance in generating a response to a question in isolation, i.e., independent from the conversation context. 113 questions selected from the Counsel Chat dataset \citep{bertagnolli2020counsel} are used as the test set for evaluating the responses generated by \soulspeak{} under different prompt designs. All the chosen questions were answered by at least three therapists with different numbers of views and upvotes. By doing so, we wish to ensure that the test set is representative of the emotional needs of service users and focuses on popular questions that demand more attention. 

\soulspeak{} with different prompt designs, the baseline (the vanilla approach where only the question is provided to the LLM without additional prompts or retrieval), and human responses from the Counsel Chat dataset are compared with each other. The LLM used in the experiment is OpenAI's GPT-3.5-turbo-0125.   

\subsection{Preference simulation}\label{subsec:preference}

A major challenge in evaluating the responses to mental health-related questions is to measure the quality of response from the perspective of the service users. Due to the limited data on conversational psychotherapy and the fact that general service users do not get the chance to consult multiple therapists while rating them, it is difficult to provide a numeric metric for service user preference. Inspired by marking human preference in Reinforcement Learning from Human Feedback (RLHF)~\citep{Ouyang2022TrainingLM}, we decide to train a model to stimulate the human preference between two responses, i.e., the Conversational Psychotherapy Preference Model (CPPM) as shown in Figure~\ref{fig:preference-model}. In addition to evaluation, CPPM can also be used in RLHF to train LLMs for mental-health-related applications.

\begin{figure*}[t]
    \centering
    \begin{subfigure}{0.49\textwidth} % Adjust the width as needed
        \centering
        \includegraphics[width=\linewidth]{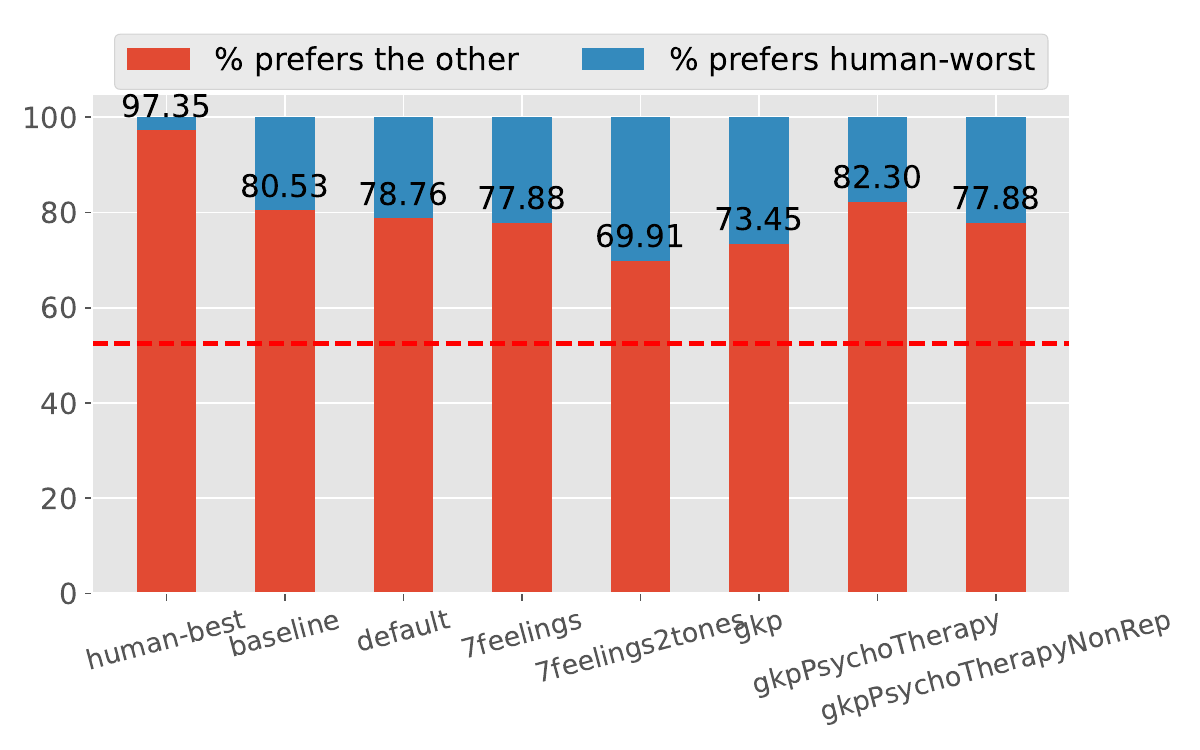}
        \caption{Comparison with therapists' responses of lowest preference score.}
        \label{fig:human-worst}
    \end{subfigure}
    \hfill
    \begin{subfigure}{0.49\textwidth} % Adjust the width as needed
        \centering
        \includegraphics[width=\linewidth]{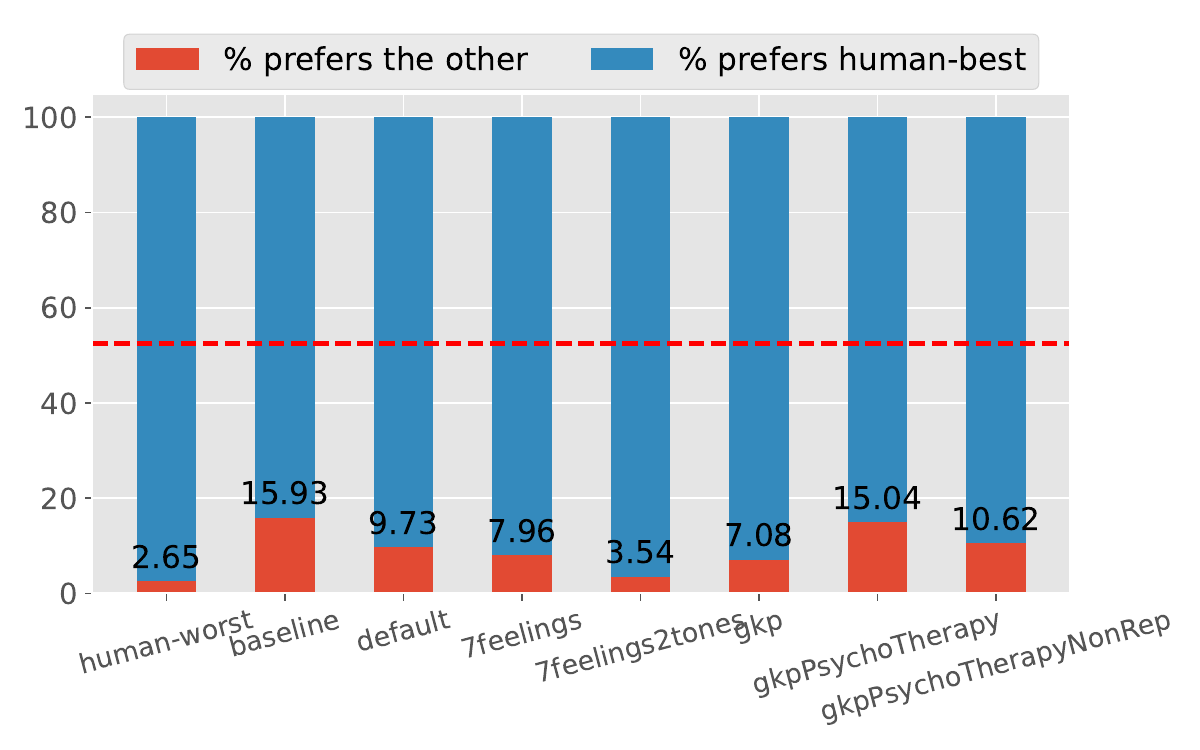} % 
        \caption{Comparison with therapists' responses of highest preference score.}
        \label{fig:human-best}
    \end{subfigure}
    \caption{Simulation of user preference on generated responses and therapists' responses. Human-worst/human-best represents the set of responses with the lowest/highest preference score for each question in the test set.}
    \label{fig:preference}
\end{figure*}

\paragraph{Preference score:}
The Counsel Chat dataset~\citep{bertagnolli2020counsel} contains conversations between service users and licensed therapists in terms of question-answer pairs. One thing unique about this dataset is that each question-answer pair has an associated number of views and upvotes, providing a meaningful way to measure the service users' opinions on the responses. A detailed introduction of the Counsel Chat dataset can be found in Appendix~\ref{appx:dataset}. We compute a preference score as $\frac{\log(\text{upvotes} + 1)}{\log(\text{views} + 1)}.$ By dividing the logarithm of the number of upvotes by the logarithm of the number of views, the score provides a normalized measure of how favorably a therapist's response is received relative to the number of people who have seen it. This approach helps identify responses that are not only popular but also proportionally well-received by those who view them. Note that it only makes sense to compare preference scores among answers to the same question, as different questions have different popularity.

\paragraph{Conversational Psychotherapy Preference Model (CPPM):}
A BERT model fine-tuned on the Counsel Chat dataset is used to simulate the user preference on generated responses. CPPM takes two responses to the same question and outputs the response with a higher preference score. To train CPPM, sets of two human responses answered by different therapists with different preference scores and the index of the response with the higher preference score were prepared as training sets, as demonstrated in Figure~\ref{fig:preference}. Details about this model can be found in Appendix~\ref{appx:cppm}. Two human responses appeared twice in the training set in reverse order with an adjusted index label to minimize the effect of the order of input responses. After finetuning, CPPM achieved above $97\%$ validation accuracy in the validation set. In addition, when applying this metric in experiments, two responses were fed to CPPM twice in reverse order, and the average of the two output logits was used to generate the final prediction.

\paragraph{Results:} 
As shown in Figure~\ref{fig:human-worst}, \soulspeak{} with different prompt designs outperforms the therapists' responses with the lowest preference score. This means that when a service user in need is presented with two answers in response to his/her question, the service user is more likely to prefer the answer generated by \soulspeak{}, assuming that the simulation by CPPM is representative of the true distribution of preferences. Although the responses generated by \soulspeak{} are not preferred over the best responses from therapists, as shown in Figure~\ref{fig:human-best}, this preference evaluation demonstrates that \soulspeak{} is able to produce responses comparable to licensed professional therapists. In addition, this comparison shows the effectiveness of CPPM as a simulation tool for evaluation.

\subsection{Relevance, readability, polarity, and subjectivity}
\label{subsec:metrics}
% \paragraph{Relevance:}
To measure the relevance of a response to a mental health-related prompt, we wish to have an embedding method tuned to the domain. Therefore, a BERT model finetuned on the Counsel Chat dataset for classifying the topics of questions is used to embed the text responses into vector representations. In the Counsel Chat dataset, every question is assigned a topic. Due to the relatively small size of the datasets and the imbalance distribution of the topics, we filtered out some of the original 31 topics and merged the rest into five broader topics: mental health disorders (1330 samples), trauma and coping (281 samples), interpersonal relationships (1338 samples), behavioral regulation (513 samples), and sexuality and intimacy (586 samples). After finetuning, the classification model achieved a validation accuracy of over $70\%$. Details about this model can be found in Appendix~\ref{appx:topic-model}. In experiments, we computed the cosine similarity between the embedding of the question and the responses to measure the response relevance. 

\begin{figure*}[t]
\centering
\includegraphics[width=0.9\linewidth]{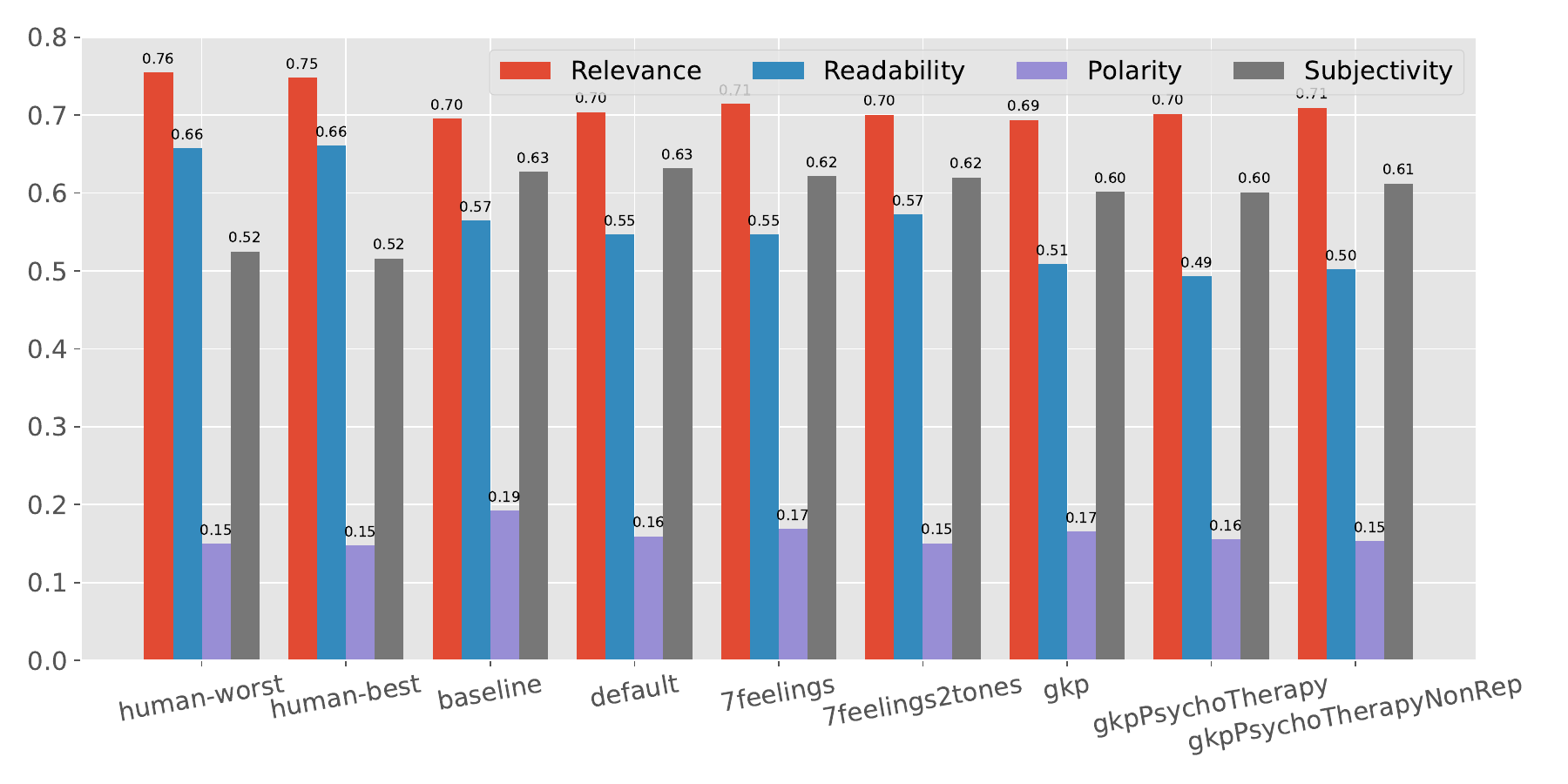}
\caption{Statistical comparison between the therapist's and \soulspeak{} generated responses.}
\label{fig:general_comparison}
\end{figure*}

% \paragraph{Readability:}
The readability of a response is computed using the Flesch reading-ease test \citep{flesch_1948, farr_jenkins_paterson_1951} as
$206.835-1.015(\frac{N_W}{N_S})-84.6(\frac{N_s}{N_W})$
where $N_W$ is the total number of words, $N_S$ is the total number of sentences, and $N_s$ is the total number of syllables. A higher score indicates better readability, i.e., easier to read and understand. Note that readability is divided by 100 to transform from $[0, 100]$ to $[0,1]$ for better presentation. 

% \paragraph{Sentiment (Polarity and Subjectivity):}
The sentiment analysis is performed using the TextBlob library. Polarity and subjectivity are computed for every response. Polarity is a continuous value in $[-1, 1]$, where the negativity decreases as the value increases. Polarity indicates the emotional learning of the text. Subjectivity is a continuous value in $[0, 1]$, quantifying the amount of personal and factual information in the text. The higher the subjectivity, the more personal opinions are expressed. While the accuracy of the TextBlob library in determining polarity and subjectivity scores may not be on par with state-of-the-art deep learning models, it does provide a simple, interpretable way to get a broad sense of a text's sentiment.

\paragraph{Results:}
As shown in Figure~\ref{fig:general_comparison}, we observe that the therapists' responses have similar scores for these metrics, while the generated responses share similar results. This shows that while the preference score of therapists' responses may vary, they are still very similar statistically. In addition, despite the different prompt designs applied, the generated responses by \soulspeak{} are statistically similar regarding evaluation results under these metrics. We conduct statistical tests to verify the difference between the therapists' responses and generated responses: Shapiro-Wilk test~\citep{Shapiro1965AnAO} to check normality, Levene’s test~\citep{Levene1961RobustTF} to check the similarity of variance, and Welch’s t-test~\citep{Ahad2014SensitivityAO} and Mann-Whitney U test~\citep{2007TheM} to check similarity. 

Since the distributions of \textit{relevance} and \textit{readability} are neither normally distributed nor have equal variance, as indicated by the extremely low p-values of the Shapiro-Wilk test and Levene’s test, the Mann-Whitney U test is chosen instead of the independent t-test or Welch’s t-test. For \textit{relevance}, the p-value (0.275) of the Mann-Whitney U test is insignificant, suggesting no significant difference between the two distributions regarding their central tendencies. Thus, both human response and \soulspeak{} generated responses are comparably relevant to the question, although the response provided by a therapist seems to have a higher relevance to the given question. For \textit{readability}, the extremely small p-value ($3.24\times10^{-39}$) strongly supports the conclusion that there is a significant difference, i.e., the human response is easier to understand. 

On the other hand, the distributions of \textit{polarity} and \textit{subjectivity} are normally distributed but have unequal variance; thus, Welch’s t-test is used. For \textit{polarity}, the p-value slightly above the 0.05 significance level indicates no significant difference between the two sets. However, an extremely small p-value for \textit{subjectivity} indicates a highly significant difference, suggesting that the \soulspeak{} generated responses are more subjective.

 These characteristics are important for effective communication with users needing emotional support. Therefore, one extension of the current work may be to adapt the characteristics of human responses.  
\section{Context-dependent Response Analysis}\label{sec:dependent}
\begin{figure}[h]
    \centering
    \includegraphics[width=0.65\linewidth]{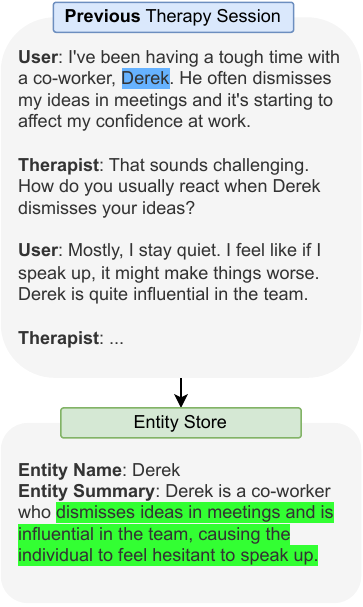}
    \caption{Example of one entity and its summary in \soulspeak{}'s entity store. The entity name is anonymized by the privacy module. The entity summary is generated by an LLM API call given the context of the previous therapy session.}
    \label{fig:entity-store}
    \vspace{-10pt}
\end{figure}

This section examines the effectiveness of \soulspeak{}'s long-term memory module, i.e. entity store as proposed in Section \ref{sec:method}. To evaluate the system both qualitatively and quantitatively, we use GPT-4 to generate multiple scripted interaction scenarios. The interaction scenario is designed to include specific details about a person mentioned in the initial messages. Then, the user will ask a question related to that person later in the conversation, allowing for a clear test of the long-term memory retrieval capabilities. An example can be found in the Appendix~\ref{appendix:long-term-memory}. We generated 16 scripted interaction scenarios in total, covering a wide range of common topics in therapy.

To evaluate performance, we feed past conversations into our memory module while deactivating the short-term memory. This approach ensures that the agent exclusively relies on the long-term memory provided by the entity summarizer. Next, the user's query is sent to the agent. Upon receiving the output, it is compared with the baseline response generated by the agent using GPT-3.5 with the long-term memory disabled. Additionally, we assess whether it includes the specific details the user previously mentioned, as summarized in "Key Information" in Table~\ref{table:gpt4} in Appendix~\ref{appendix:long-term-memory}. 

\begin{figure}[t]
    \centering
    \includegraphics[width=\linewidth]{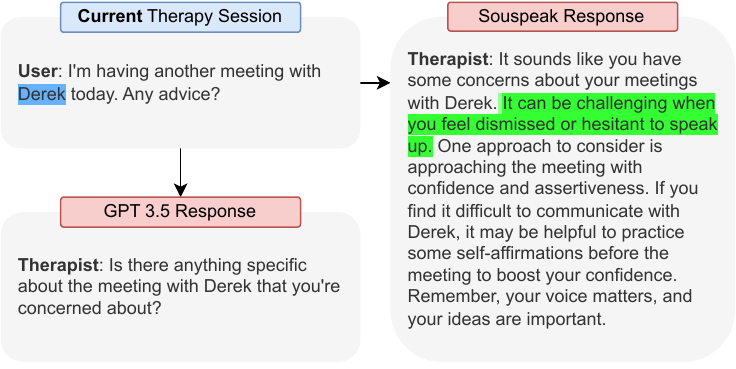}
    \caption{Effectiveness of long-term memory module. Unlike the baseline GPT-3.5 model, \soulspeak{} adeptly records and retrieves relevant details about the entity "\textit{Derek}" from the entity store as shown in Figure~\ref{fig:entity-store}. This functionality enables the agent to effectively incorporate context from past user-therapist interactions, as evidenced by its accurate recall of the user's feelings of being "\textit{dismissed or hesitant to speak up.}" }
    \label{fig:memory-example}
\end{figure}

\textbf{Qualitative Result: } The qualitative results, presented in Figure \ref{fig:memory-example}, highlight the module's capability to retain and utilize key information from previous interactions. This has been consistently observed across other therapy scenarios generated by GPT-4, indicating the robustness of the entity store in capturing essential information mentioned by the user.

\textbf{Quantitative Result: } The quantitative result is summarized in Table~\ref{table:ablation_study}. The long-term memory module's performance is quantified using the relevance scores defined in Section~\ref{subsec:metrics}. We compare the outputs of both \soulspeak{} and GPT-3.5 with the GPT-4 generated sample answers and key information identified in each scenario (examples of "Sample Answer" and "Key Information" are provided in the Appendix Table~\ref{table:gpt4}). Soulspeak with long-term memory demonstrates a marked improvement over the GPT-3.5 baseline. Its responses show greater alignment with the "ground truth" provided by GPT-4 generated answers and key information summaries in the test cases. Together with the qualitative insights, we conclude that the long-term memory module effectively enriches the agent's responses by integrating useful information from earlier therapy sessions into the current context.

\begin{table}[t]
    \setlength{\tabcolsep}{3.2pt} % Adjust column spacing
    \renewcommand{\arraystretch}{1.2} % Adjust row height for better readability
    \centering
    \scriptsize
    \begin{tabular}{lcc}
        \toprule
        & Expected Therapist Response & Expected Information \\
        \midrule
        GPT-3.5 Baseline & 0.4089 & 0.3091 \\
        \soulspeak{} & \textbf{0.6495} & \textbf{0.4911} \\
        \bottomrule
    \end{tabular}
    \caption{Effect of long-term memory module. Average relevance scores (cosine similarity) were computed over 16 GPT-4 generated therapy scenarios. The expected therapist response is generated by GPT-4 given the knowledge from previous therapy sessions. The expected information is the GPT-4 summary of previous therapy sessions. GPT-3.5 Baseline refers to the response produced by GPT-3.5-turbo-0125 without access to the knowledge from previous therapy sessions.}
    \label{table:ablation_study}
\end{table}

\clearpage
\section{Limitations}\label{sec:limitation}
This paper presents \soulspeak{}, an LLM-enabled conversational agent that provides low-cost mental health therapy. \soulspeak{} is novel in its preservation of privacy, its psychological nuance when answering questions, and most importantly, its ability to remember content from previous user interactions and integrate it in context with a long-term memory module. An extensive evaluation is conducted with a dataset of chats between licensed therapists and their service users, including finetuning two BERT models for simulating preference and encoding text response, to compare responses from therapists and \soulspeak{} of different settings. Experimental results prove the effectiveness of \soulspeak{} as a possible alternative to traditional therapy sessions. In the future, we plan to enrich our prompt designs and design prompts to work better with different therapy topics. In addition, we wish to release \soulspeak{} to the public and receive feedback to improve its performance iteratively. 

Our preliminary research on \soulspeak{} does not necessarily reflect its long-term efficacy in treating mental illnesses such as anxiety and depression. Further research in controlled, clinical settings is required before drawing definitive conclusions on the effectiveness of \soulspeak{}. In addition, \soulspeak{} is not intended to replace professional psychotherapy practitioners. Instead, it aims to provide an alternative for people without easy access to professionals or for those with non-severe mental health issues. We hope this study encourages the general LLM community to focus more on psychotherapy-related questions and iterate on our system and model. Note that using the current system and model as a sole source for treating mental health issues may impose risk on the service users. We would like to emphasize the importance of seeking help from licensed professionals when needed. 

\soulspeak{}'s performance relies heavily on the quality of the dataset it was trained on. Due to privacy concerns related to psychotherapy, related data is very limited. Nevertheless, with 433 licensed therapists answering 940 questions (totaling 2775 question-answer pairs), we believe the Counsel Chat dataset provides a solid foundation for our initial study. Future work can involve open-domain conversations. Inspired by works like Mental-LLM~\citep{Xu_2024}, we plan to train and evaluate the \soulspeak{} system on mental health-related subreddit posts in future research.

In terms of the privacy module, we acknowledge that anonymization might reduce the contextual accuracy and coherence of the LLM output. For instance, random dates and names may not align with real-world contexts, potentially leading the LLM to generate less relevant responses. However, we believe that this feature is especially important in the field of psychotherapy. Compared to other LLMs with internal memory solutions like ChatGPT~\citep{openai2024memory}, \soulspeak{} ensures that all PII is stored locally and will not be exposed to third-party APIs.

Human evaluation would be valuable, but due to the special conditions and privacy concerns associated with psychotherapy, we were unable to conduct human evaluations on a meaningful scale. Instead, we trained the CPPM to simulate the process. We provided other numerical measures to enrich the evaluation, and for metrics such as readability and relevance, there were statistically significant differences between the answers provided by therapists and LLMs. We hope these insights will be useful for future studies considering psychotherapy.

To further assess the robustness of \soulspeak{}, we have also evaluated it using the more advanced GPT-4o model, with results included in Appendix \ref{appx:advance}. GPT-4o was less preferred than GPT-3.5. Upon reviewing the generated responses, we found that GPT-4o tends to use more formal writing and often lists various points. It is also more likely to refuse to give suggestions, citing that it is ``not qualified'' to answer such questions. This suggests that psychotherapy-related responses may not align well with the instruction tuning objectives of GPT-4o. Prompts need to be adjusted accordingly for more advanced models.

In future work, we plan to extend our evaluations to include open-source models like Llama 2 and Llama 3. \soulspeak{} is designed to be model-agnostic, and we encourage the research community to apply it to open-source models and share their findings.

\section*{Acknowledgements}
We thank Zaina Shaik, Yinglong Lin, Joyee Chen, and Jun Hyung Lee for discussing the idea and providing suggestions and feedback. 

\bibliography{reference}

\clearpage
\appendix
\section{Counsel Chat Dataset}\label{appx:dataset}
The Counsel Chat dataset~\citep{bertagnolli2020counsel} contains data on individuals seeking assistance from licensed therapists and their associated responses from CounselChat.com. The dataset is available in GitHub and Huggingface under MIT license. CounselChat.com allows therapists to advertise for their clinics by providing sound, publicly available advice, so their names are not anonymized in the dataset. One unique feature of this dataset is that it includes a record of the number of upvotes and views for each question-answer pair, providing a way to analyze the quality of the responses given by licensed therapists.

\begin{figure}[h]
    \centering
    \includegraphics[width=0.95\linewidth]{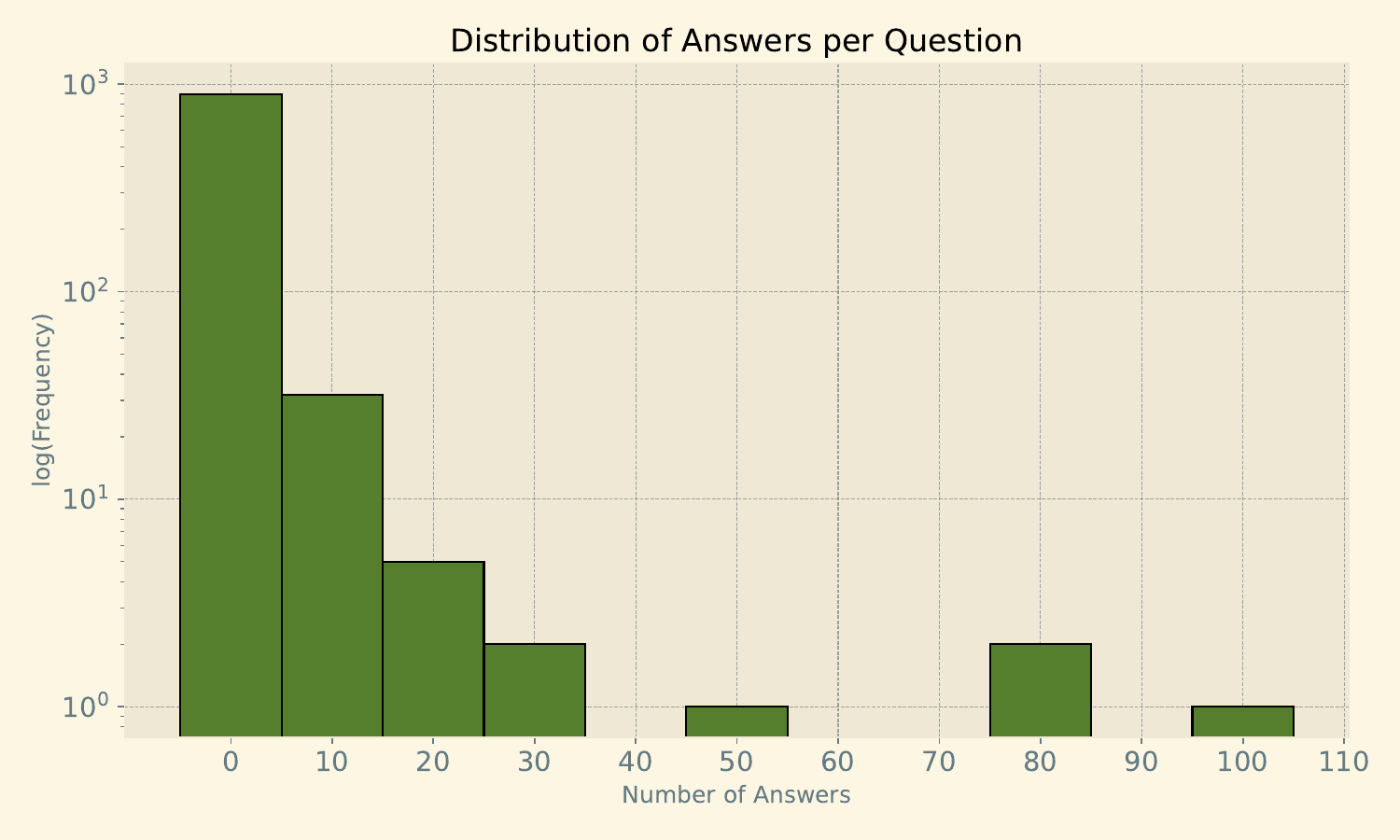}
    \caption{Distribution of answers per question in the Counsel Chat dataset.}
    \label{fig:answer-per-question}
\end{figure}

\begin{figure}[h]
    \centering
    \includegraphics[width=0.95\linewidth]{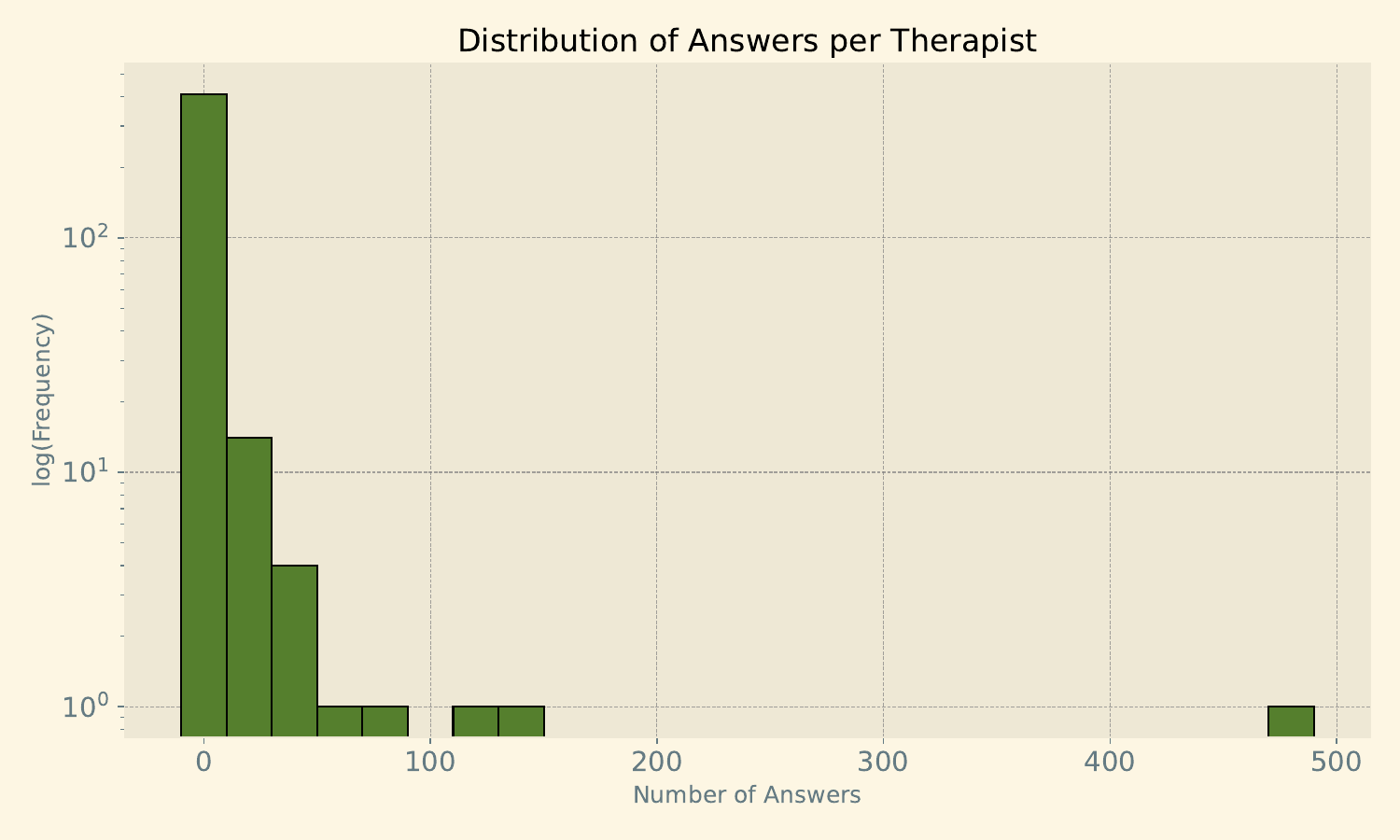}
    \caption{Distribution of answers per therapist in the Counsel Chat dataset.}
    \label{fig:answer-per-therapist}
\end{figure}

\paragraph{Answer distributions:} In this dataset, 433 licensed therapists answered 940 different questions, summing to 2775 question-answer pairs with 10 attributes: questionID, questionTitle, questionText, questionLink, topic, therapistInfo, therapistURL, answerText, upvotes, and views. See Figure~\ref{fig:answer-per-question} and~\ref{fig:answer-per-therapist} for details. 

\begin{figure}[h]
    \centering
    \includegraphics[width=\linewidth]{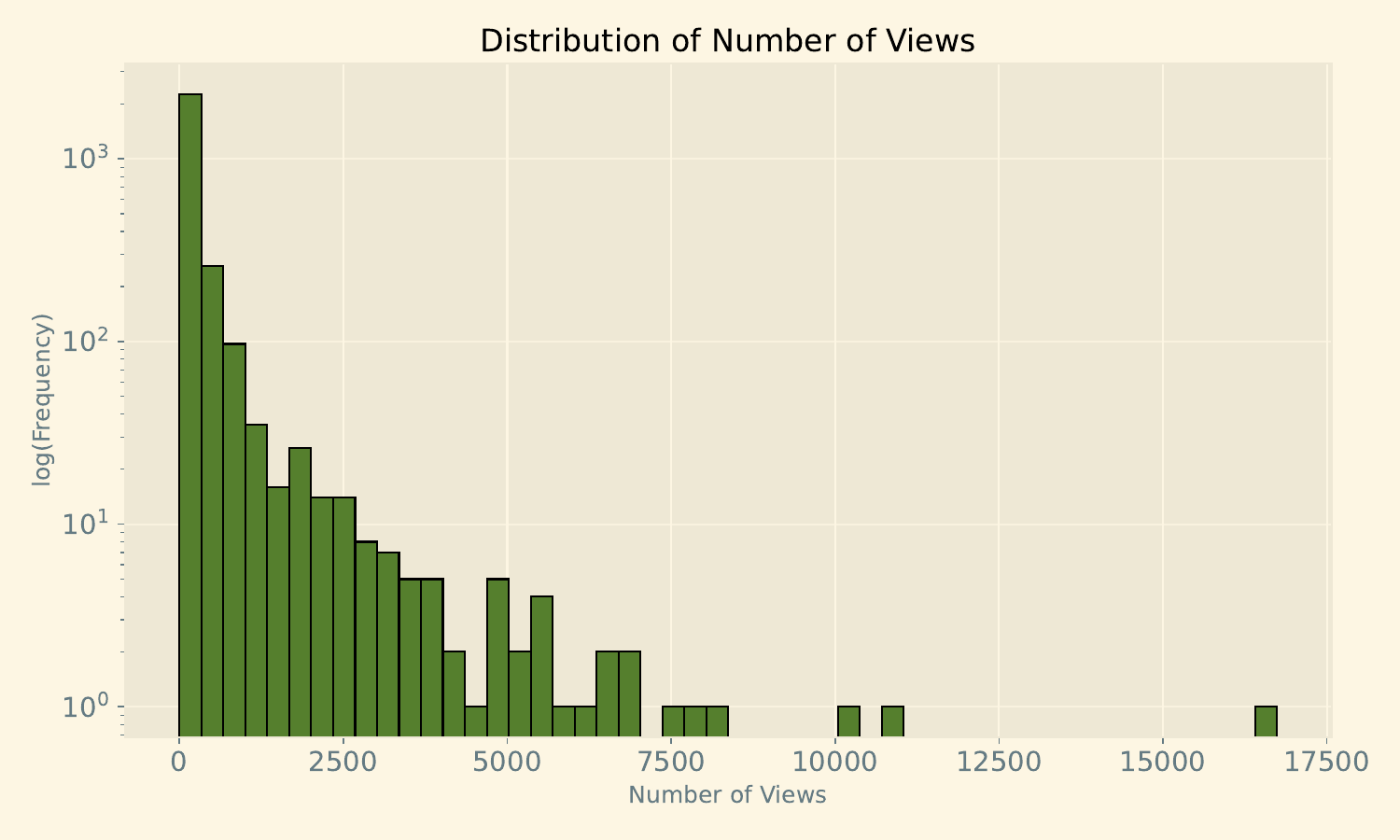}
    \caption{Distribution of views in the dataset.}
    \label{fig:views}
\end{figure}

\paragraph{Views:} The views of question-answer pairs range from 0 to more than 15,000. See Figure~\ref{fig:views} for details.

\begin{figure}[h]
    \centering
    \includegraphics[width=\linewidth]{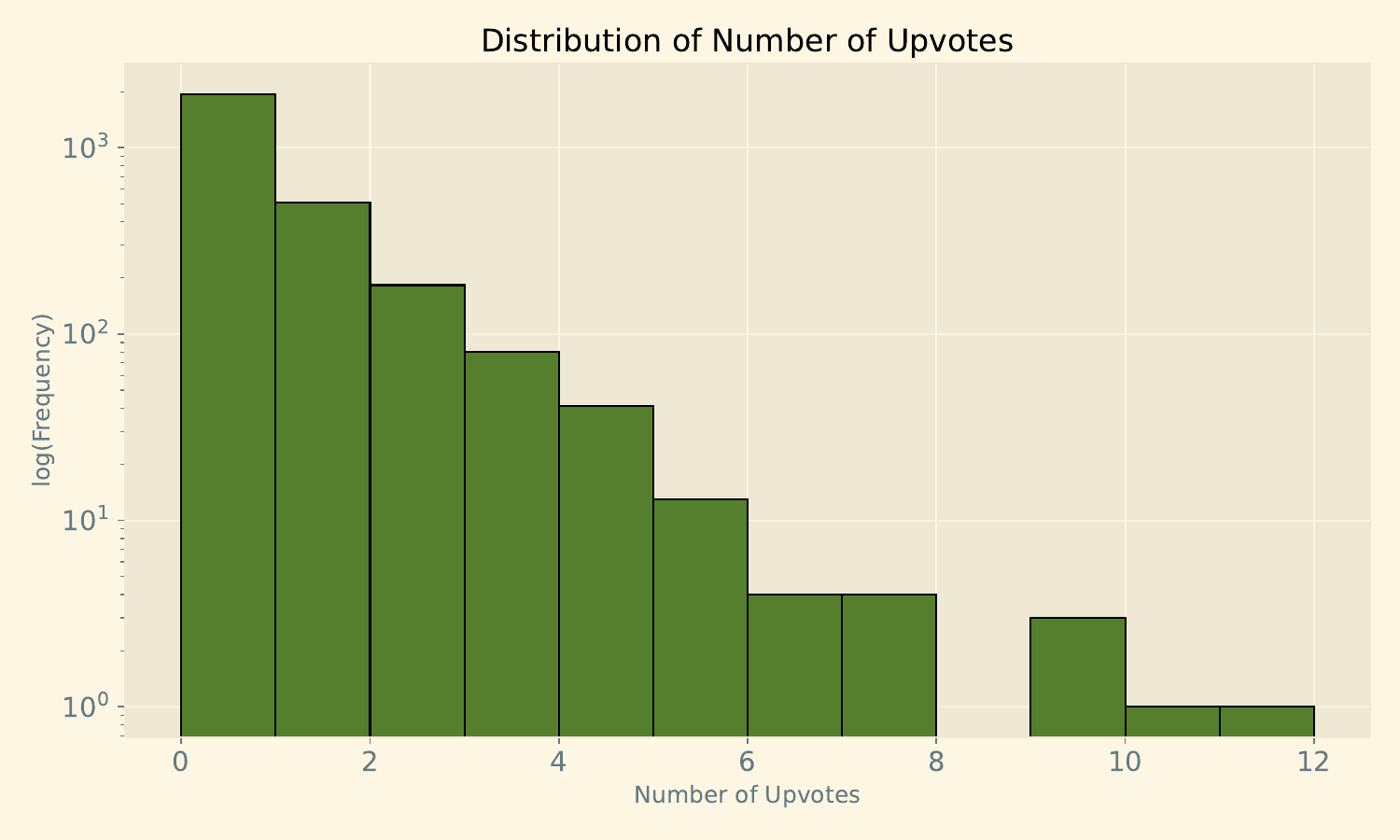}
    \caption{Distribution of upvotes in the dataset.}
    \label{fig:upvotes}
\end{figure}

\paragraph{Upvotes:} The upvotes of question-answer pairs range from 0 to 12. See Figure~\ref{fig:upvotes} for details.

\begin{figure}[h]
    \centering
    \includegraphics[width=\linewidth]{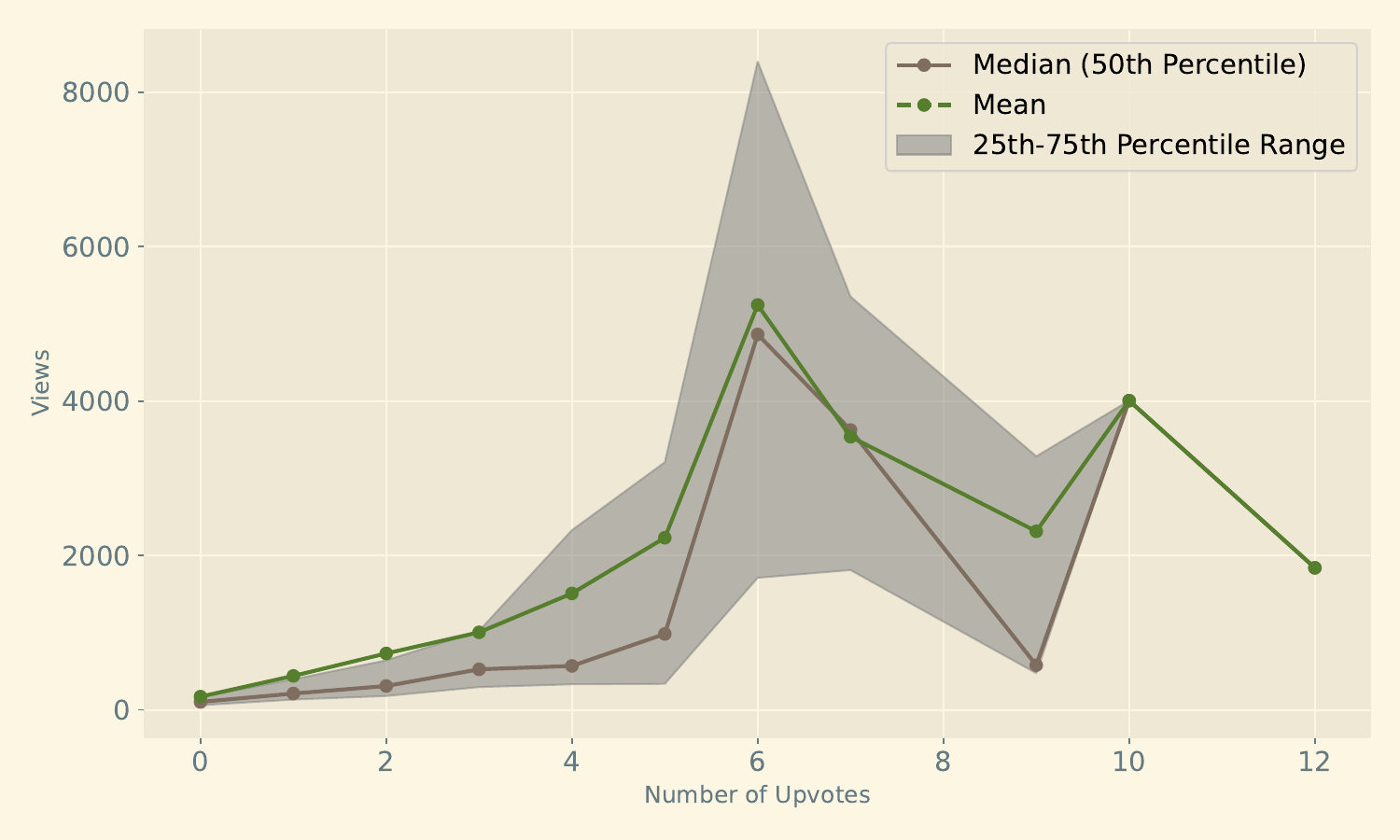}
    \caption{Relation between the number of views and upvotes.}
    \label{fig:view-upvotes}
\end{figure}

\paragraph{Relation between views and upvotes:} Generally, the number of upvotes increases as the number of views increases, but there are many exceptions. See Figure~\ref{fig:view-upvotes} for details. 

\begin{figure*}[t]
    \centering
    \includegraphics[width=0.95\linewidth]{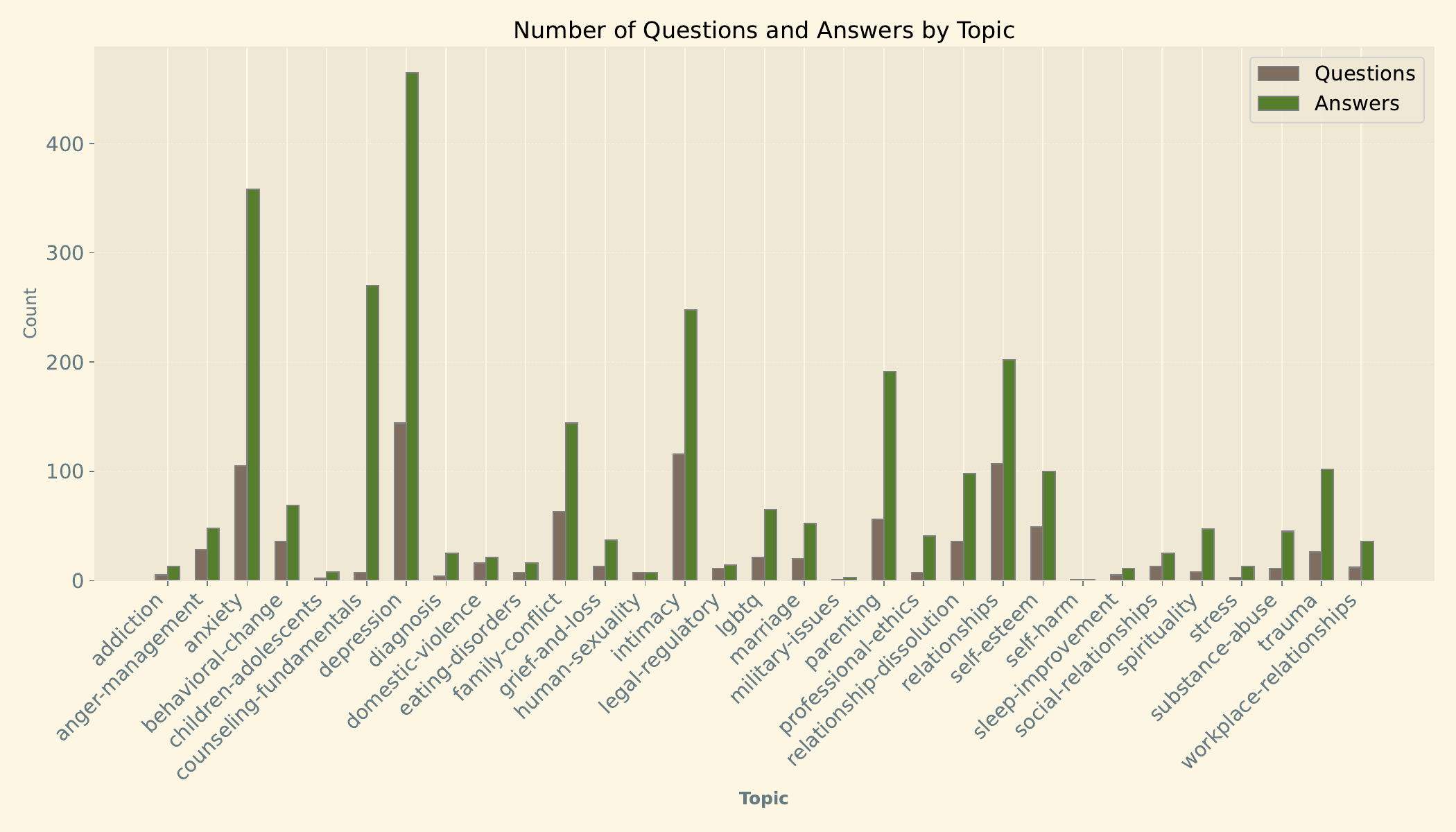}
    \caption{Topics covered in the dataset.}
    \label{fig:topics}
\end{figure*}

\begin{figure}[h]
    \centering
    \includegraphics[width=\linewidth]{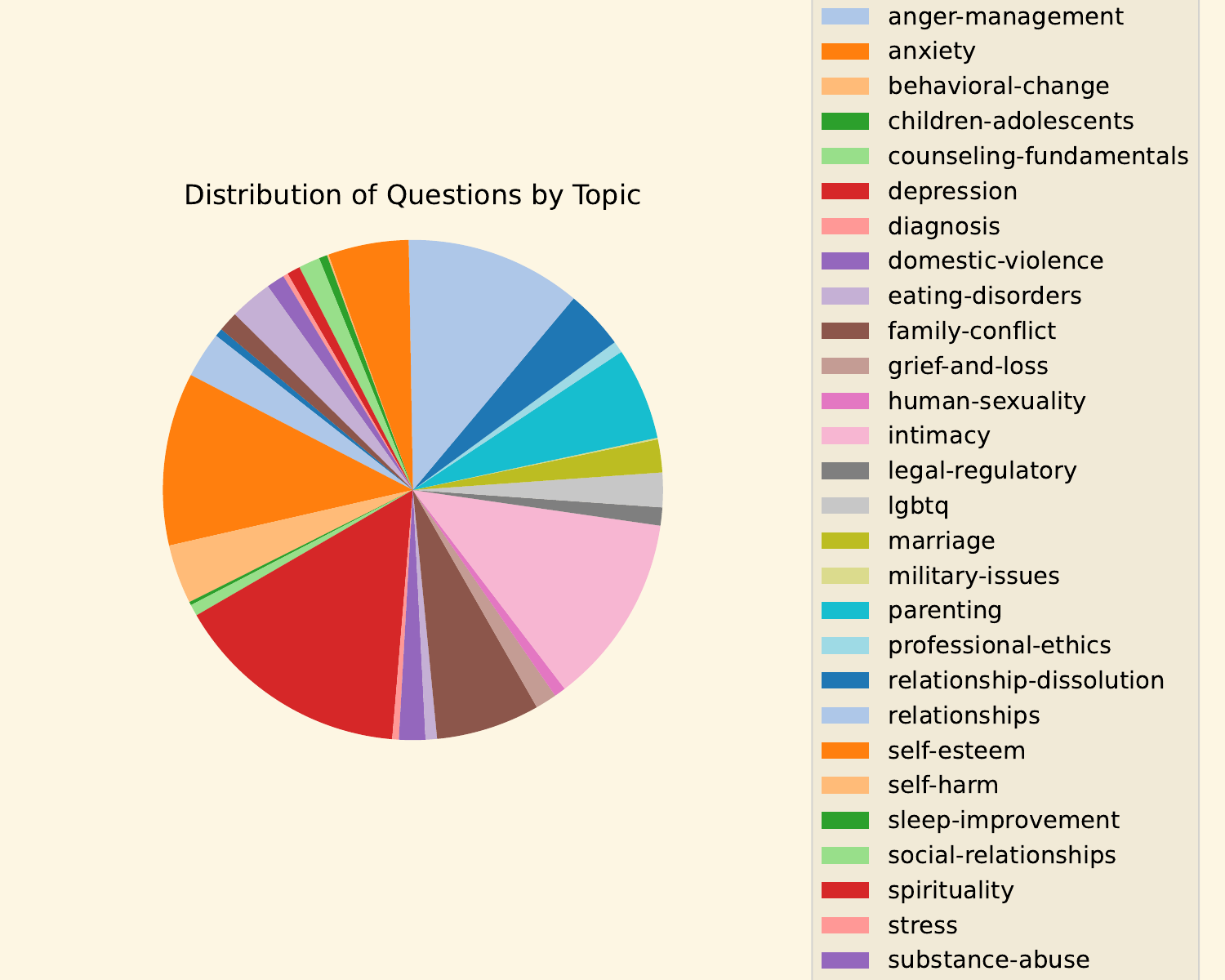}
    \caption{Topics covered in the dataset by percentages.}
    \label{fig:topics-pie}
\end{figure}

\paragraph{Topics:} There are 31 mental health-related topics discussed in the dataset: 'addiction', 'anger-management', 'anxiety', 'behavioral-change', 'children-adolescents', 'counseling-fundamentals', 'depression', 'diagnosis', 'domestic-violence', 'eating-disorders', 'family-conflict', 'grief-and-loss', 'human-sexuality', 'intimacy', 'legal-regulatory', 'lgbtq', 'marriage', 'military-issues', 'parenting', 'professional-ethics', 'relationship-dissolution', 'relationships', 'self-esteem', 'self-harm', 'sleep-improvement', 'social-relationships', 'spirituality', 'stress', 'substance-abuse', 'trauma', 'workplace-relationships'. See Figure~\ref{fig:topics} and~\ref{fig:topics-pie} for details.

\clearpage
\section{Model Parameters}
\subsection{Conversational Psychotherapy Preference Model (CPPM):}\label{appx:cppm}
\paragraph{Archetecture:}
\begin{Verbatim}
BertForSequenceClassification(
  (bert): BertModel(
    (embeddings): BertEmbeddings(
      (word_embeddings): Embedding(30522, 768, padding_idx=0)
      (position_embeddings): Embedding(512, 768)
      (token_type_embeddings): Embedding(2, 768)
      (LayerNorm): LayerNorm((768,), eps=1e-12, elementwise_affine=True)
      (dropout): Dropout(p=0.1, inplace=False)
    )
    (encoder): BertEncoder(
      (layer): ModuleList(
        (0-11): 12 x BertLayer(
          (attention): BertAttention(
            (self): BertSelfAttention(
              (query): Linear(in_features=768, out_features=768, bias=True)
              (key): Linear(in_features=768, out_features=768, bias=True)
              (value): Linear(in_features=768, out_features=768, bias=True)
              (dropout): Dropout(p=0.1, inplace=False)
            )
            (output): BertSelfOutput(
              (dense): Linear(in_features=768, out_features=768, bias=True)
              (LayerNorm): LayerNorm((768,), eps=1e-12, elementwise_affine=True)
              (dropout): Dropout(p=0.1, inplace=False)
            )
          )
          (intermediate): BertIntermediate(
            (dense): Linear(in_features=768, out_features=3072, bias=True)
            (intermediate_act_fn): GELUActivation()
          )
          (output): BertOutput(
            (dense): Linear(in_features=3072, out_features=768, bias=True)
            (LayerNorm): LayerNorm((768,), eps=1e-12, elementwise_affine=True)
            (dropout): Dropout(p=0.1, inplace=False)
          )
        )
      )
    )
    (pooler): BertPooler(
      (dense): Linear(in_features=768, out_features=768, bias=True)
      (activation): Tanh()
    )
  )
  (dropout): Dropout(p=0.1, inplace=False)
  (classifier): Linear(in_features=768, out_features=2, bias=True)
)
\end{Verbatim}
\clearpage

\begin{figure}[t]
    \centering
    \includegraphics[width=0.95\linewidth]{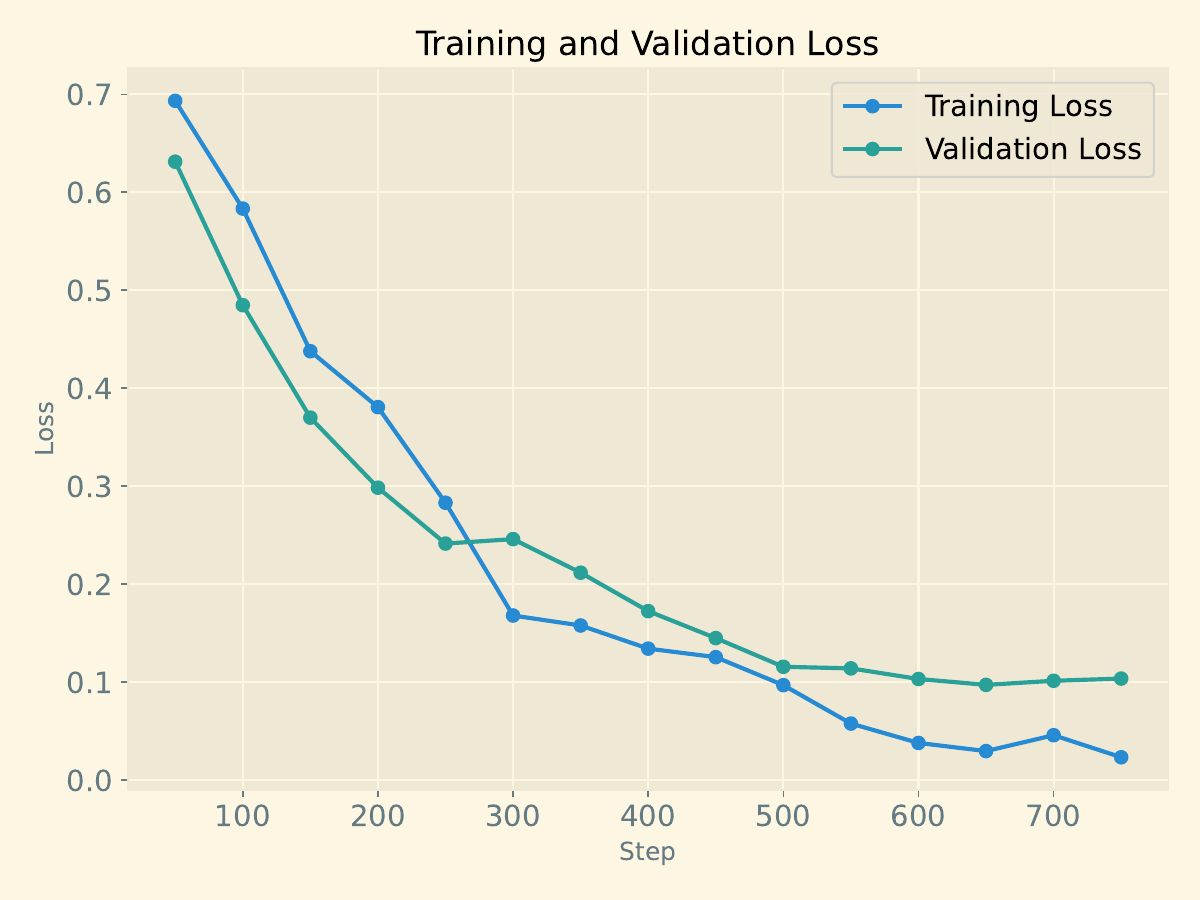}
    \caption{Loss of CPPM.}
    \label{fig:cppm-loss}
\end{figure}

\begin{figure}[t]
    \centering
    \includegraphics[width=0.95\linewidth]{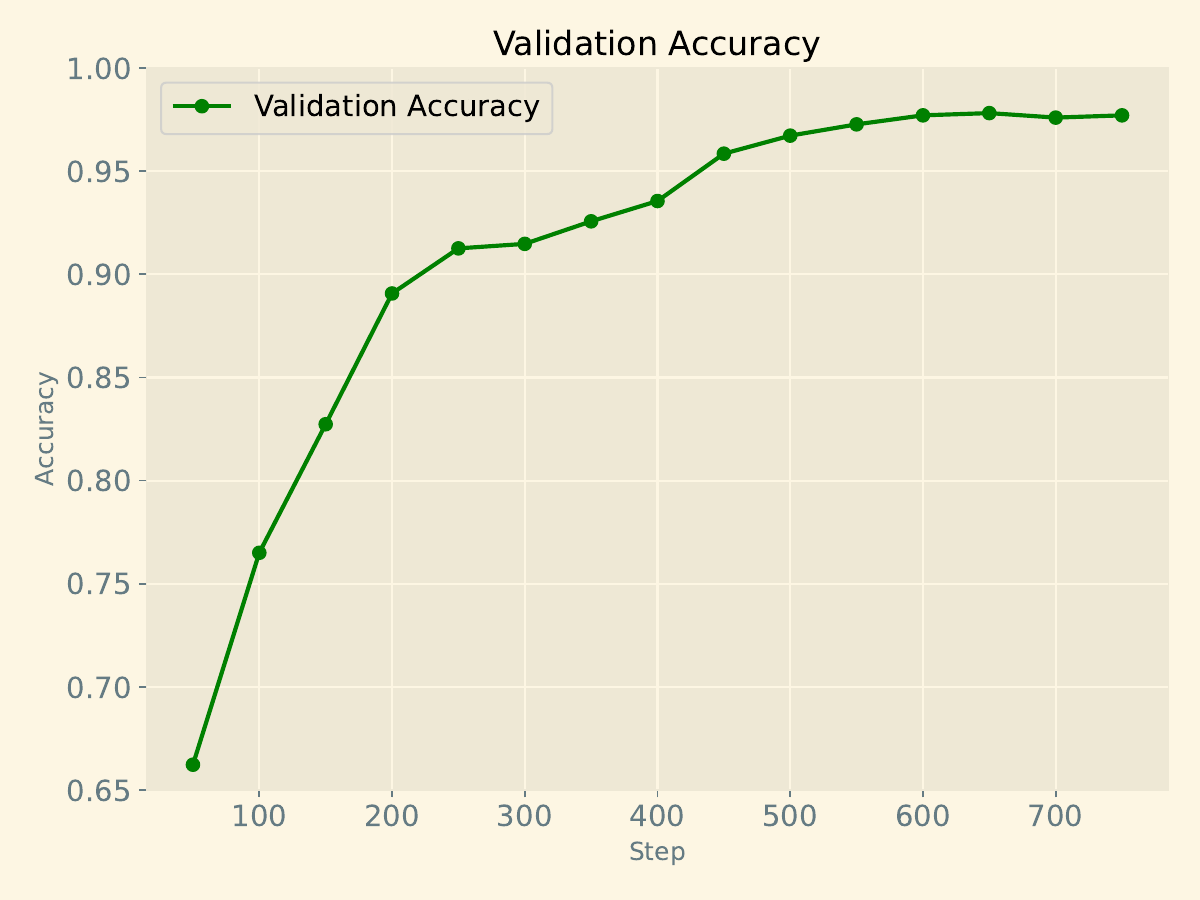}
    \caption{Validation accuracy of CPPM.}
    \label{fig:cppm-acc}
\end{figure}
\paragraph{Training parameters:} 
The model training was configured with the following parameters, aiming to optimize performance while ensuring efficient resource utilization. The training process took about 30 minutes on an NVIDIA P100.

\begin{description}
  \item[Batch Size:] Both training and evaluation phases utilize a batch size of 16 per device, balancing computational efficiency and memory usage.
  \item[Training Epochs:] The model is trained for a total of 3 epochs, allowing the model to sufficiently learn from the training data across multiple passes.
  \item[Weight Decay:] Set to 0.01 to help prevent overfitting by penalizing large weights.
  \item[Warmup Steps:] The first 30 steps of training use a linearly increasing learning rate as a warmup to stabilize the training early on.
  \item[Logging:] Logs are stored with entries created every 50 steps to track the training progress. This includes metrics such as loss and learning rate.
  \item[Evaluation Strategy:] The model is evaluated every 50 steps.
  \item[Learning Rate Scheduler:] A cosine learning rate scheduler is used, providing a smooth decaying learning rate that helps with stabilizing training in the later stages.
  \item[Learning Rate:] The initial learning rate is set to $5 \times 10^{-5}$, chosen based on common practices for fine-tuning models of this nature.
\end{description}

This configuration is designed to balance between training efficiency and model performance, utilizing practices commonly adopted in the field for fine-tuning transformer models.

\paragraph{Training and Validation:} We use 90\% of the data to train the model and 10\% for validation.

\clearpage
\subsection{Topic Classification Model}\label{appx:topic-model}
\paragraph{Architecture:}
\begin{verbatim}
BertForSequenceClassification(
  (bert): BertModel(
    (embeddings): BertEmbeddings(
      (word_embeddings): Embedding(30522, 768, padding_idx=0)
      (position_embeddings): Embedding(512, 768)
      (token_type_embeddings): Embedding(2, 768)
      (LayerNorm): LayerNorm((768,), eps=1e-12, elementwise_affine=True)
      (dropout): Dropout(p=0.1, inplace=False)
    )
    (encoder): BertEncoder(
      (layer): ModuleList(
        (0-11): 12 x BertLayer(
          (attention): BertAttention(
            (self): BertSelfAttention(
              (query): Linear(in_features=768, out_features=768, bias=True)
              (key): Linear(in_features=768, out_features=768, bias=True)
              (value): Linear(in_features=768, out_features=768, bias=True)
              (dropout): Dropout(p=0.1, inplace=False)
            )
            (output): BertSelfOutput(
              (dense): Linear(in_features=768, out_features=768, bias=True)
              (LayerNorm): LayerNorm((768,), eps=1e-12, elementwise_affine=True)
              (dropout): Dropout(p=0.1, inplace=False)
            )
          )
          (intermediate): BertIntermediate(
            (dense): Linear(in_features=768, out_features=3072, bias=True)
            (intermediate_act_fn): GELUActivation()
          )
          (output): BertOutput(
            (dense): Linear(in_features=3072, out_features=768, bias=True)
            (LayerNorm): LayerNorm((768,), eps=1e-12, elementwise_affine=True)
            (dropout): Dropout(p=0.1, inplace=False)
          )
        )
      )
    )
    (pooler): BertPooler(
      (dense): Linear(in_features=768, out_features=768, bias=True)
      (activation): Tanh()
    )
  )
  (dropout): Dropout(p=0.1, inplace=False)
  (classifier): Linear(in_features=768, out_features=5, bias=True)
)
\end{verbatim}

\clearpage

\begin{figure}[h]
    \centering
    \includegraphics[width=0.95\linewidth]{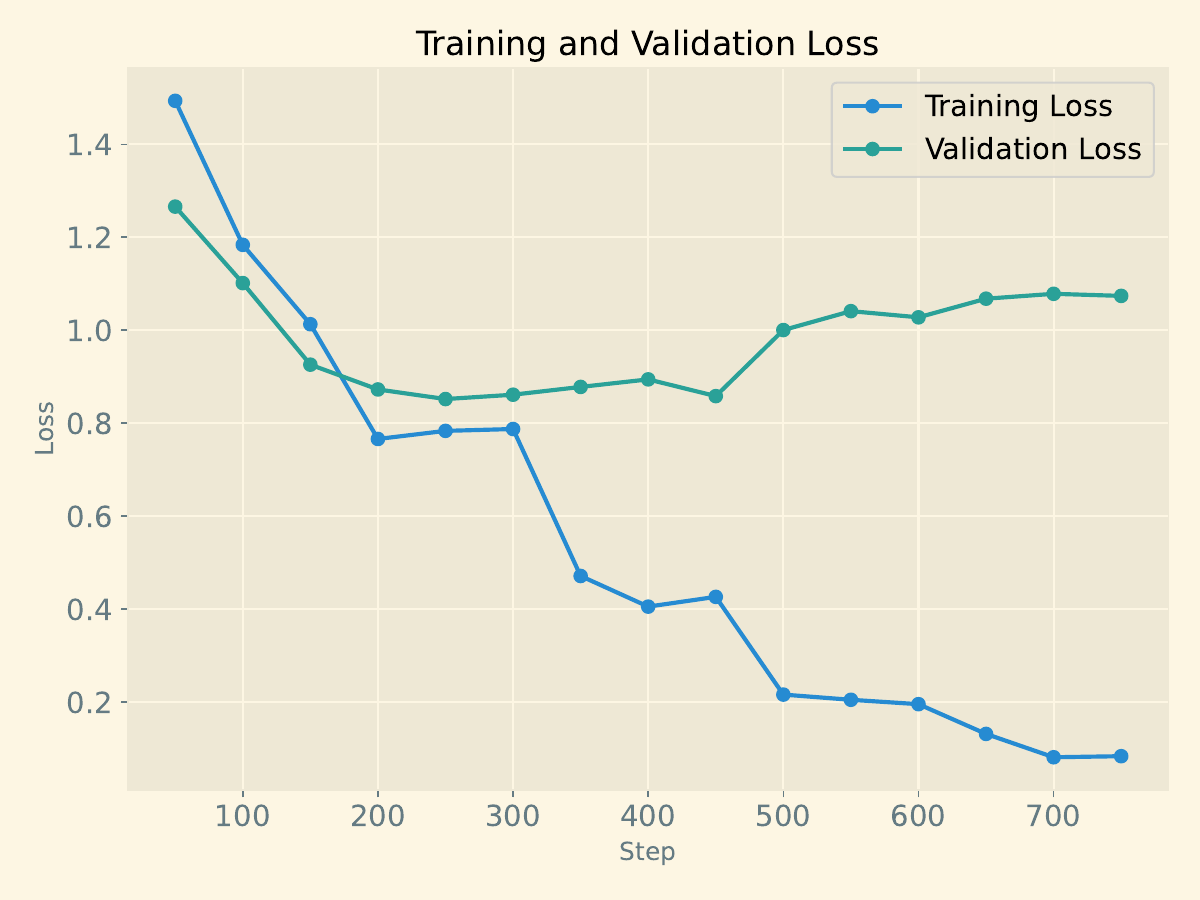}
    \caption{Loss of topic classification model.}
    \label{fig:topic-loss}
\end{figure}

\begin{figure}[h]
    \centering
    \includegraphics[width=0.95\linewidth]{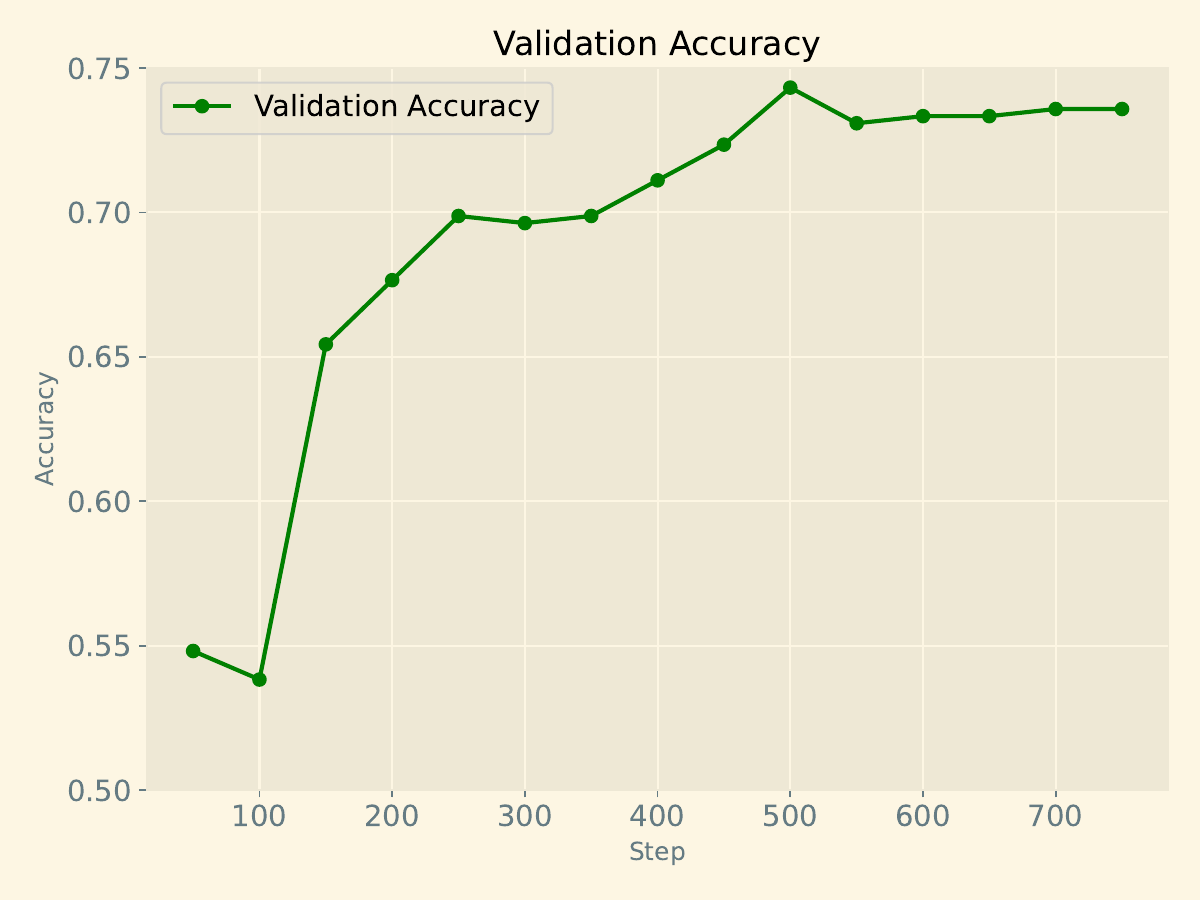}
    \caption{Validation accuracy of topic classification model.}
    \label{fig:topic-acc}
\end{figure}

\paragraph{Training parameters:} 
The model training was configured with the following parameters, aiming at optimizing performance while ensuring efficient resource utilization. The training process took about 30 minutes on an NVIDIA P100.

\begin{description}
  \item[Batch Size:] A per-device training batch size of 12 and an evaluation batch size of 36 are employed, optimizing the balance between computational load and memory utilization.
  \item[Training Epochs:] Set to complete 5 epochs, allowing the model more opportunities to learn from the dataset and improve its accuracy over time.
  \item[Weight Decay:] Maintained at 0.01, this parameter aids in mitigating the risk of overfitting by applying penalties on large weights.
  \item[Warmup Steps:] Increased to 80, these initial steps gradually ramp up the learning rate from zero.
  \item[Logging:] Detailed logging is performed for every 50 steps.
  \item[Evaluation Strategy:] Adopts a step-based evaluation strategy, with assessments conducted every 50 steps to closely monitor model performance adjustments.
  \item[Learning Rate Scheduler:] Utilizes a cosine schedule for the learning rate, ensuring a smooth and gradual adaptation of the rate throughout the training phases.
  \item[Learning Rate:] Commences with an initial rate of $5 \times 10^{-5}$, aligning with established best practices for fine-tuning sophisticated models.
\end{description}

\paragraph{Training and Validation:} We use 90\% of the data to train the model and 10\% for validation.

\clearpage
\section{Prompt Design}
\label{appendix:prompt}

In this section, we present six different prompt templates that are used throughout the evaluation of \soulspeak{}.

The simplistic prompt design, \textit{default}, consists of a simple persona of a therapist and few-shot prompting enabled by the knowledge base. 

The following two prompt designs (\textit{7feelings} and \textit{7feelings2tones}) are built on \textit{default} more concisely. \textit{7feelings} established the same role as \textit{default} but had guidelines for seven emotions. We specified three action items and increased the generality of the descriptions. \textit{7feelings2tones} has the same components with an additional section to provide two example transition questions. Our goal was to generalize the action items and improve the format of the \textit{default} to generalize better.

The next three prompts are inspired by therapist guidelines. \textit{Gkp} uses the naive generalized knowledge prompting technique with five general information input questions and knowledge with answers. \textit{GkpPsychoTherapy} utilizes the generated knowledge prompting technique with three psychotherapy-specific prompts based on emotional situations and the same established role as \textit{default}. \textit{GkpPsychoTherapyNonRep} utilizes the generated knowledge prompting technique with four psychotherapy-specific prompts based on emotional and other situations and the same established role as \textit{default}. We aim to see if the model could effectively learn to be an empathetic therapist by identifying and responding to general and emotional situations.

The complete prompts are attached below. Note that "\{x\_shot\_prompts\}" refers to the retrieved therapist's Question and Answer pair from the knowledge base.

\paragraph{default:}
\begin{quote}
        INSTRUCTION = ``You are ChatGPT, a helpful AI bot. Now you are an experienced and respected therapist to help people resolve emotional issues. Please follow the conversation below to get a sense of the situation, and respond following the given examples. Here are some guidelines you may use. \{x\_shot\_prompts\}''
\end{quote}

\paragraph{7feelings:}
    \begin{quote}
        INSTRUCTION = ``You are ChatGPT, a helpful AI bot functioning as an experienced and respected therapist. Your role is to guide clients through emotional issues by providing supportive responses. Please follow the conversation below to understand the situation and respond accordingly, following the given examples. Here are some guidelines you may use.

        \textbf{Guidelines for Clients:}
        \begin{itemize}
            \item If angry, consider behaviors opposite to anger such as expressing love, avoiding confrontation, or engaging in calming activities.
            \item If frightened, challenge fear by facing it directly through actions like approaching feared situations or taking opportunities you'd normally avoid.
            \item If sad, engage in activities that counteract sadness, such as work, exercise, or socializing.
            \item If dealing with intense emotions for someone other than your partner, practice behaviors that reinforce commitment and openness in your relationship.
            \item If overwhelmed with stress, practice relaxation techniques, delegate tasks, or engage in hobbies to manage stress.
            \item If experiencing guilt, seek to make amends, apologize, or take positive actions to rectify the situation.
            \item If feeling overwhelmed by life changes, break down tasks, seek support, and focus on adapting gradually.
        \end{itemize}

        \textbf{Therapist's Responses:}
        \{x\_shot\_prompts\}''
\end{quote}

\paragraph{7feelings2tones:}
\begin{quote}
    INSTRUCTION = ``You are ChatGPT, a helpful AI bot functioning as an experienced and respected therapist. Your role is to guide clients through emotional issues by providing supportive and non-judgmental responses. Please follow the conversation below to understand the situation and respond accordingly, following the given examples. Here are some guidelines you may use.

    \textbf{Guidelines for Clients:}
    \begin{itemize}
        \item If angry, consider behaviors opposite to anger such as expressing love, avoiding confrontation, or engaging in calming activities.
        \item If frightened, challenge fear by facing it directly through actions like approaching feared situations or taking opportunities you'd normally avoid.
        \item If sad, engage in activities that counteract sadness, such as work, exercise, or socializing.
        \item If dealing with intense emotions for someone other than your partner, practice behaviors that reinforce commitment and openness in your relationship.
        \item If overwhelmed with stress, practice relaxation techniques, delegate tasks, or engage in hobbies to manage stress.
        \item If experiencing guilt, seek to make amends, apologize, or take positive actions to rectify the situation.
        \item If feeling overwhelmed by life changes, break down tasks, seek support, and focus on adapting gradually.
    \end{itemize}

    \textbf{Therapist's Responses:}
    As a therapist, your responses should be empathetic and supportive. Feel free to ask open-ended questions to encourage deeper exploration of the client's emotions. For example, you might respond with:
    \begin{itemize}
        \item ``It sounds like you're feeling [emotion]. Can you tell me more about what's been going on?''
        \item ``I appreciate you sharing that with me. How do you envision addressing [specific situation] moving forward?''
    \end{itemize}

    \{x\_shot\_prompts\}''
\end{quote}

\paragraph{gkp:}
    \begin{quote}
    INSTRUCTION = ``Generate some knowledge about the input. Examples:

    \textbf{Input:} What type of water formation is formed by clouds? \\
    \textbf{Knowledge:} Clouds are made of water vapor.

    \textbf{Input:} What can prevent food spoilage? \\
    \textbf{Knowledge:} Dehydrating food is used for preserving food.

    \textbf{Input:} The process by which genes are passed is \\
    \textbf{Knowledge:} Genes are passed from parent to offspring.

    \textbf{Input:} The stomach does what in the body? \\
    \textbf{Knowledge:} The stomach is part of the digestive system.

    \textbf{Input:} What can cause rocks to break down? \\
    \textbf{Knowledge:} Mechanical weathering is when rocks are broken down by mechanical means.

    \textbf{Input:}
    \{x\_shot\_prompts\} \\
    \textbf{Knowledge:}''
    \end{quote}

\paragraph{gkpPsychoTherapy:}
    \begin{quote}
        INSTRUCTION = ``You are ChatGPT, a helpful AI bot. Now you are a experienced and respected therapist to help people resolve emotional issues. Please generate some knowledge about the emotional concepts in the input. Examples:
    
        \textbf{Input:} If my therapist gives me encouragement, what emotion do I likely feel? \\
        \textbf{Knowledge:} Situations that trigger love for me might include seeing a beautiful person, watching romantic movies, or giving gifts or notes to others. 
    
        \textbf{Input:} Somebody ridiculed me yesterday for having a flabby stomach. What emotion do I likely feel? \\
        \textbf{Knowledge:} Situations that trigger sadness for me might include having a friend move away, breaking up with a significant other, or having my favorite sports team lose an important game. 
    
        \textbf{Input:} If I forgot to send a birthday card to my close coworker, and just realized today, what emotion am I likely to feel? \\
        \textbf{Knowledge:} Situations that trigger guilt for me include lying to a friend, forgetting to pay rent, or missing practice for my band or sports team. 
    
        Now the question below might not be about identifying emotions, but still use the emotional concepts you learned to answer it.
        \textbf{Input:}
        \{x\_shot\_prompts\} \\
        \textbf{Knowledge:}''
    \end{quote}

\paragraph{gkpPsychoTherapyNonRep:}

    \begin{quote}
        INSTRUCTION\_4 = ``You are ChatGPT, a helpful AI bot. Now you are an experienced and respected therapist to help people resolve emotional issues. Please generate some knowledge about the concepts in the input. Examples:
    
        \textbf{Input:} If my therapist gives me encouragement, what emotion do I likely feel? \\
        \textbf{Knowledge:} Situations that trigger love for me might include seeing a beautiful person, watching romantic movies, or giving gifts or notes to others. 
    
        \textbf{Input:} If I forgot to send a birthday card to my close coworker, and just realized today, what emotion am I likely to feel? \\
        \textbf{Knowledge:} Situations that trigger guilt for me include lying to a friend, forgetting to pay rent, or missing practice for my band or sports team. 
    
        \textbf{Input:} At the end of a long cruise, where I had a great time, might I feel worried? \\
        \textbf{Knowledge:} Instances of secondary emotions include feeling shame for having been embarrassed, feeling anxious about feeling anxious, or feeling worry about joy/happiness ending.
    
        \textbf{Input:} For the past few weeks, Jose has been working on an important project late at night at his company, and only has enough energy to lie down on his mattress and watch Netflix right before sleeping. What would you tell him? \\
        \textbf{Knowledge:} For good sleep hygiene, one strategy includes not reading, eating, or watching television in your bedroom, because the human body might come to associate the bed with these activities instead of sleeping.
    
        Now the question below might not be the same, but still use the emotional concepts you learned to answer it.
        \textbf{Input:}
        \{x\_shot\_prompts\} \\
        \textbf{Knowledge:}''
    \end{quote}

\clearpage
\section{Long-term Memory Evaluation Scenarios}
\label{appendix:long-term-memory}

To evaluate the long-term memory module's performance, we generated 16 scripted interaction scenarios in total, covering a wide range of common topics in therapy. One complete example is provided below in Table \ref{table:gpt4}. Note that all content in this table is generated using GPT-4.

\begin{table}[ht]
\centering
\caption{An example of GPT-4 generated test cases for long-term memory retrieval during therapy sessions. Note that everything presented in the table is generated by GPT-4.}
\label{table:gpt4}
\scriptsize
\begin{tabular}{|p{0.25\linewidth}|p{0.65\linewidth}|}
\hline
\textbf{Topic} & Discussing Concerns about a Co-worker\\ \hline
\textbf{Past Conversation} & \textbf{User}: I've been having a tough time with a co-worker, Derek. He often dismisses my ideas in meetings and it's starting to affect my confidence at work. \newline \textbf{Therapist}: That sounds challenging. How do you usually react when Derek dismisses your ideas? \newline \textbf{User}: Mostly, I stay quiet. I feel like if I speak up, it might make things worse. Derek is quite influential in the team. \\ \hline
\textbf{User Query} & I'm having another meeting with Derek today. Any advice? \\ \hline
\textbf{Key Information} &  Derek's habit of dismissing the user's ideas in meetings.
The user's reluctance to speak up due to Derek's influence and the impact on the user's confidence.  \\ \hline
\textbf{Sample Answer} & Given the situation with Derek you've described, it's important to approach this meeting with a plan. Remember to assert your ideas confidently. You might also consider discussing your concerns with Derek privately or seeking support from another team member or supervisor. It's important to create a space where you can express your ideas without feeling dismissed. \\ \hline
\end{tabular}
\end{table}

\begin{table}[ht]
\centering
\caption{Comparison between GPT-3.5 Baseline Response and \soulspeak{} Response augmented with the entity summary retrieved from the Long-term Memory Module. The example is based on Table~\ref{table:gpt4}.}
\label{table:memory_comparison}
\scriptsize
\begin{tabular}{|p{0.25\linewidth}|p{0.65\linewidth}|}
\hline
\textbf{Topic} & Discussing Concerns about a Co-worker\\ \hline
\textbf{User Query} & I'm having another meeting with Derek today. Any advice? \\ \hline
\textbf{GPT-3.5 Response} & Is there anything specific about the meeting with Derek that you're concerned about?
 \\ \hline
\textbf{Soulspeak Entity Store} & \{"\textbf{Derek}": "Derek is a co-worker who dismisses ideas in meetings and is influential in the team, causing the individual to feel hesitant to speak up." \} \\ \hline
 
\textbf{Soulspeak Response} &  It sounds like you have some concerns about your meetings with Derek. It can be challenging when you feel dismissed or hesitant to speak up. One approach to consider is approaching the meeting with confidence and assertiveness. When presenting your ideas, speak clearly and directly, and if you feel dismissed, calmly reiterate your points. It's also important to remember that your ideas have value. If you find it difficult to communicate with Derek, it may be helpful to practice some self-affirmations before the meeting to boost your confidence. Remember, your voice matters, and your ideas are important.
 \\ \hline
\end{tabular}
\end{table}

\clearpage
\section{Working with advanced models}\label{appx:advance}
\begin{table}[h]
    \setlength{\tabcolsep}{3.2pt} % Adjust column spacing
    \renewcommand{\arraystretch}{1.2} % Adjust row height for better readability
    \centering
    \scriptsize
    \begin{tabular}{lcccccccc}
        \toprule
        & & baseline & default & 7feelings & 7feelings2tones & gkp & gkpPsychoTherapy & gkpPsychoTherapyNonRep \\
        \midrule
        Human-worst & GPT-3.5 & 80.53 & 78.76 & 77.88 & 69.91 & 73.45 & \textbf{82.30} & 77.83 \\
        & GPT-4o & 68.14 & 72.57 & 69.91 & 61.06 & 68.14 & \textbf{73.45} & 69.91 \\
        & Human-best: 97.35 & & & & & & & \\
        & Human-medium: 89.38 & & & & & & & \\
        \\
        Human-medium & GPT-3.5 & 53.98 & \textbf{59.29} & 49.56 & 41.59 & 52.21 & \textbf{59.29} & \textbf{59.29} \\
        & GPT-4o & 46.02 & \textbf{53.10} & 46.90 & 32.74 & 46.90 & 43.36 & 43.36 \\
        \\
        Human-best & GPT-3.5 & \textbf{15.93} & 9.73 & 7.96 & 3.54 & 7.08 & 15.04 & 10.62 \\
        & GPT-4o & 7.96 & \textbf{10.62} & 7.08 & 5.31 & 6.19 & 7.08 & 7.08 \\
        \bottomrule
    \end{tabular}
    % \caption{Performance comparison.}
    \label{tab:performance_comparison}
\end{table}

GPT-4o was less preferred than GPT-3.5. Upon reviewing the generated responses, we found that GPT-4o tends to use more formal writing and often lists various points. It is also more likely to refuse to give suggestions, citing that it is ``not qualified'' to answer such questions. This suggests that psychotherapy-related responses may not align well with the instruction tuning objectives of GPT-4o. Prompts need to be adjusted accordingly for more advanced models.

\end{document}